\documentclass{article}

\usepackage[nonatbib,final]{neurips_2022}

\usepackage[utf8]{inputenc} % allow utf-8 input
\usepackage[T1]{fontenc}    % use 8-bit T1 fonts
\usepackage{hyperref}       % hyperlinks
\usepackage{url}            % simple URL typesetting
\usepackage{booktabs}       % professional-quality tables
\usepackage{amsfonts}       % blackboard math symbols
\usepackage{nicefrac}       % compact symbols for 1/2, etc.
\usepackage{microtype}      % microtypography
\usepackage{xcolor}         % colors

\usepackage{amsmath,amssymb}
\usepackage{graphicx}
\usepackage{colortbl}
\usepackage[numbers]{natbib}
\usepackage{siunitx}

\newcommand{\greyrule}{\arrayrulecolor{black!30}\midrule\arrayrulecolor{black}}

\hypersetup{
    colorlinks,
    linkcolor={red!50!black},
    citecolor={blue!50!black},
    urlcolor={blue!80!black}
}

\title{Non-Markovian Reward Modelling from Trajectory Labels via Interpretable Multiple Instance Learning}

\author{
  Joseph Early
  \thanks{Equal contribution}
  \ 
  \thanks{University of Southampton, United Kingdom; \texttt{\{J.A.Early,C.Evers,sdr1\}@soton.ac.uk}} 
  \
  $^\mathsection$
 \And
   Tom Bewley
   \footnotemark[1]
   \ 
   \thanks{University of Bristol, United Kingdom; \texttt{tom.bewley@bristol.ac.uk}}
   \
   \thanks{The Alan Turing Institute, United Kingdom}
 \And
   Christine Evers
   \footnotemark[2]
 \And
   Sarvapali Ramchurn
   \footnotemark[2]
   \ 
   \footnotemark[4]
}

\begin{document}

\maketitle

\begin{abstract}
% We generalise the problem of reward modelling (RM) for reinforcement learning (RL) to handle non-Markovian rewards. Existing work assumes that human evaluators observe each step in a trajectory independently when providing feedback on agent behaviour. In this work, we remove this assumption, extending RM to include hidden state information that captures temporal dependencies in human assessment of trajectories. We then show how RM can be approached as a multiple instance learning (MIL) problem, and develop new MIL models that are able to capture the time dependencies in labelled trajectories. We demonstrate on a range of RL tasks that our novel MIL models can reconstruct reward functions to a high level of accuracy, and that they provide interpretable learnt hidden information that can be used to train high-performing agent policies.
We generalise the problem of reward modelling (RM) for reinforcement learning (RL) to handle non-Markovian rewards. Existing work assumes that human evaluators observe each step in a trajectory independently when providing feedback on agent behaviour. In this work, we remove this assumption, extending RM to capture temporal dependencies in human assessment of trajectories. We show how RM can be approached as a multiple instance learning (MIL) problem, where trajectories are treated as bags with return labels, and steps within the trajectories are instances with unseen reward labels. We go on to develop new MIL models that are able to capture the time dependencies in labelled trajectories. We demonstrate on a range of RL tasks that our novel MIL models can reconstruct reward functions to a high level of accuracy, and can be used to train high-performing agent policies.
\end{abstract}

\section{Introduction}

There is growing consensus around the view that aligned and beneficial AI requires a reframing of objectives as being contingent, uncertain, and learnable via interaction with humans \citep{russell2019human}. In reinforcement learning (RL), this proposal has found one formalisation in reward modelling (RM): the inference of agent objectives from human preference information such as demonstrations, pairwise choices, approval labels, and corrections \citep{leike2018scalable}. Prior work in RM typically assumes that a human evaluates the \textit{return} (quality) of a sequential trajectory of agent behaviour by summing equal and independent \textit{reward} assessments of instantaneous states and actions, with the aim of RM being to reconstruct the underlying reward function.
%carrying no dependencies from earlier events to their assessment of later ones.
%expressed preferences are
%approximately rational
%softly optimal 
%with respect to a
%latent \textit{return}
%%latent fitness
%function on the space of trajectories -- sequences of environment states $s$ and agent actions $a$ -- and seeks to reconstruct this function from the available information  \citep{jeon2020reward}.
%Prior works commonly assume that return decomposes into a sum of \textit{rewards} over constituent state-action pairs, $R(s,a)$, and attempt to reconstruct $R$. This is a highly restrictive model of human psychology, postulating an independent and equal assessment of each pair, followed by an accurate summation. 
However, in reality the human's experience of a trajectory is likely to be temporally extended (e.g., via a video clip \citep{christiano2017deep} or real-time observation), which opens the door to dependencies between earlier events and the assessment of later ones. The independence assumption may be both psychologically unrealistic given human memory limitations \citep{kahneman2000evaluation}, and technically na\"ive given the difficulty of building complete instantaneous state representations
%for real-world tasks
\cite{kaelbling1998planning}. We thus seek to generalise RM to allow for temporal dependencies in human evaluation, by postulating \textit{hidden state} information that accumulates over a trajectory. Reconstruction of the human's preferences now requires the modelling of hidden state dynamics alongside the reward function itself.
%(see Figure \ref{fig:overview}).

In tackling this generalised problem, we identify a structural isomorphism between RM
%and the established field of multiple instance learning (MIL) \citep{carbonneau2018multiple} \N{more citations here}. In MIL, data are organised into bags of instances \textemdash{} in our work the instances are state-action pairs grouped together into trajectories (bags), where instance labels are the rewards for individual timesteps, and bag labels are the returns for the overall trajectories. Using this relationship, we propose novel MIL model architectures which use long short-term memory (LSTM) modules \citep{hochreiter1997long} to recover the hidden state dynamics, and learn instance-level reward predictions from return-labelled trajectories of arbitrary length. We use these MIL models not only for RM, but also for integration into RL agent training, demonstrating they are able to boost performance by uncovering the hidden state information in NMRDPs.
(specifically from trajectory return labels) and the established field of multiple instance learning (MIL) \citep{carbonneau2018multiple}. Trajectories are recast as \textit{bags} and constituent state-action pairs as \textit{instances}, which collectively contribute to labels provided at the bag level by interacting in potentially complex ways. This mapping inspires a range of novel MIL model architectures that use long short-term memory (LSTM) modules \citep{hochreiter1997long} to recover the hidden state dynamics, and learn instance-level reward predictions from return-labelled trajectories of arbitrary length. In experiments with synthetic oracle labels, we show that our MIL RM models can accurately reconstruct ground truth hidden states and reward functions for non-Markovian tasks, and can be straightforwardly integrated into RL agent training to achieve performance matching, or even exceeding, that of agents with direct access to true hidden states and rewards. We then apply interpretability analysis to understand what the models have learnt.

Our contributions are as follows:
\begin{enumerate}
    \item We generalise RM to handle \textit{non-Markovian} rewards that depend on hidden features of the environment or the psychology of the human evaluator in addition to visible states/actions.
    \item We identify a structural connection between RM and MIL, creating the opportunity to transfer concepts and methods between the two fields.
    \item We propose novel LSTM-based MIL models for this generalised RM problem, and develop interpretability techniques for understanding and verifying the learnt reward functions.
    \item We compare our proposed models to existing MIL baselines on five non-Markovian tasks, evaluating return prediction, reward prediction, robustness to label noise, and interpretability.
    \item We demonstrate that the hidden state and reward predictions of our MIL RM models can be used by RL agents to solve non-Markovian tasks.
\end{enumerate}

% \begin{figure}[t!]
%     \centering
%     \includegraphics[width=\textwidth]{figures/overview.pdf}
%     \caption{An overview of our approach. Starting with labelled trajectories, we first learn a reward model that is able to make reward predictions for environment states and also produces latent representations capturing the hidden information in the trajectories that occurs due to temporal dependencies. These latent representations are then utilised in RL agent training to produce better performing agent policies. \N{Placeholder figure.}}
%     \label{fig:overview}
% \end{figure}

The remainder of this work is as follows. Section \ref{sec:background} discusses related work in RM and MIL, Section \ref{sec:methodology} gives a formal problem definition and describes our MIL-inspired methodology, and Section \ref{sec:results} presents our experiments and results. We discuss key findings in Section \ref{sec:discussion}, and Section \ref{sec:conclusion} concludes. All of our source code is available in a public repository.\footnote[1]{\url{https://github.com/JAEarly/MIL-for-Non-Markovian-Reward-Modelling}}

\section{Background and Related Work}
\label{sec:background}

% Using paragraph instead of subsection for compactness!
\paragraph{Reward Modelling}

RM \citep{leike2018scalable} aims to infer a reward function from revealed human preference information such as demonstrations \citep{ng2000algorithms}, pairwise choices \citep{christiano2017deep}, corrections \cite{bajcsy2017learning}, good/bad/neutral labels \citep{reddy2020learning}, or combinations thereof \citep{jeon2020reward}. Most prior work assumes a human evaluates a trajectory by summing independent rewards for each state-action pair, but in practice their experience is likely to be temporally extended (e.g., via a video clip), creating the opportunity for dependencies to emerge between earlier events and the assessment of later ones. As noted by  \citet{chan2020impacts} and \citet{bewley2022interpretable}, such dependencies may arise from cognitive biases such as anchoring, prospect bias, and the peak-end rule \citep{kahneman2000evaluation}, but they could equally reflect rational drivers of human preferences not captured by the state representation.
%It is telling that \citet{christiano2017deep} felt it necessary to urge experimental participants to \textit{``only decide on events you actually witness in the clip"}; this could be seen as an implicit acceptance that the Markov assumption is strained in practice.
Some efforts have been made to model temporal dependencies, such as a discrete psychological mode which evolves over consecutive queries about hypothetical trajectories \citep{basu2019active}, or a monotonic bias towards more recently-viewed timesteps due to human memory limitations \citep{lee2021bpref}. Elsewhere, \citet{shah2020interactive} use human demonstrations and binary approval labels to learn temporally extended task specifications in logical form. In comparison to these restricted examples, our work provides a more general approach to capturing temporal dependencies in RM.
%In recent work on RM from demonstrations, \citet{chan2021human} find that correctly modelling structured biases may actually yield \textit{improved} reward reconstruction versus using perfectly rational demonstrations.

\paragraph{Non-Markovian Rewards}
In the canonical RL problem setup of a Markov decision process (MDP), rewards depend only on the most recent state-action pair. In a non-Markovian reward decision process (NMRDP) \citep{bacchus1996rewarding}, rewards depend on the full preceding trajectory \citep{bacchus1996rewarding}. NMRDPs can be \textit{expanded} into MDPs (and thus solved by RL) by augmenting the state with a hidden state that captures all reward-relevant historical information, but this is typically not known \textit{a priori}. Data-driven approaches to learning NMRDP expansions \citep{icarte2018using} often make use of domain-specific propositions and temporal logic operators \citep{bacchus1997structured,thiebaux2006decision,toro2019learning}. Outside of the RM context, recurrent architectures such as LSTMs have been used in NMRDPs to reduce reliance on pre-specified propositions \citep{jarboui2021trajectory}. They also have a long history of use in partially observable MDPs, where dynamics are also non-Markovian \citep{bakker2001reinforcement, hausknecht2015deep, wierstra2010recurrent}.

% Online Learning of Non-Markovian Reward Models
% Joint Inference of Reward Machines and Policies for Reinforcement Learning
% Induction of subgoal automata for reinforcement learning
% Planning With Uncertain Specifications (PUnS)
% Learning non-markovian reward models in mdps
% Reinforcement Learning with Non-Markovian Rewards
% Decision-theoretic planning with non-markovian rewards
% Teaching multiple tasks to an rl agent using ltl

% \cite{neider2021advice} learn reward machines from ``advice" consisting of DFAs over hidden state alphabet which may lead to high reward.

%\cite{jarboui2021trajectory} use LSTM to learn a latent state that separates a finite number of sequential subtasks using a contrastive loss function

% \cite{guo2021edge} use LSTM as an explainability tool. post hoc identification of critical timesteps in a trajectory (closest to embedding-space LSTM).

\paragraph{Multiple Instance Learning}
In MIL \citep{carbonneau2018multiple}, datasets are structured as collections of bags $X_i \in \mathbf{X}$, each of which is comprised of instances $\{x_1^i,\ldots,x_k^i\}$ and has an associated bag-level label $Y_i$ and instance-level labels $\{y_1^i,\ldots,y_k^i\}$. The aim is to construct a model that learns solely from bag labels; instance labels are not available during training, but may be used later to evaluate instance-level predictions. The simplest MIL approaches assume that instances are independent and that the bag is unordered, but models exist for capturing various types of instance dependencies \citep{ilse2018attention, tu2019multiple, wang2018revisiting}. LSTMs have emerged as a natural architecture for modelling temporal dependencies among ordered bags, where they can be utilised to aggregate instance information into an overall bag representation. They have previously been applied to standard MIL benchmarks \citep{wang2020defense}, as well as specific problems such as Chinese painting image classification \citep{li2020multi}. As we discuss in Section \ref{sec:models}, these existing models are somewhat unsuitable for use in RM, leading us to propose our own novel model architectures.

\section{Methodology}
\label{sec:methodology}

In this section, we present the core methodology of our work. We formally define the new paradigm of non-Markovian RM (Section \ref{sec:problem_definition}), before drawing on the MIL literature to propose models that can be used to solve this generalised problem (Section \ref{sec:models}). We then go on to discuss how we can use our learnt RM models for training RL agents on non-Markovian tasks (Section \ref{sec:method_augmenting_rl}).

\subsection{Formal Definition of Non-Markovian RM}
\label{sec:problem_definition}

%the agent's action $a_t$ conditions \N{the change in the environment state $s_{t+1}\sim D(s|s_t,a_t)$}

\begin{figure}[htb!]
    \centering
    \includegraphics[width=\textwidth]{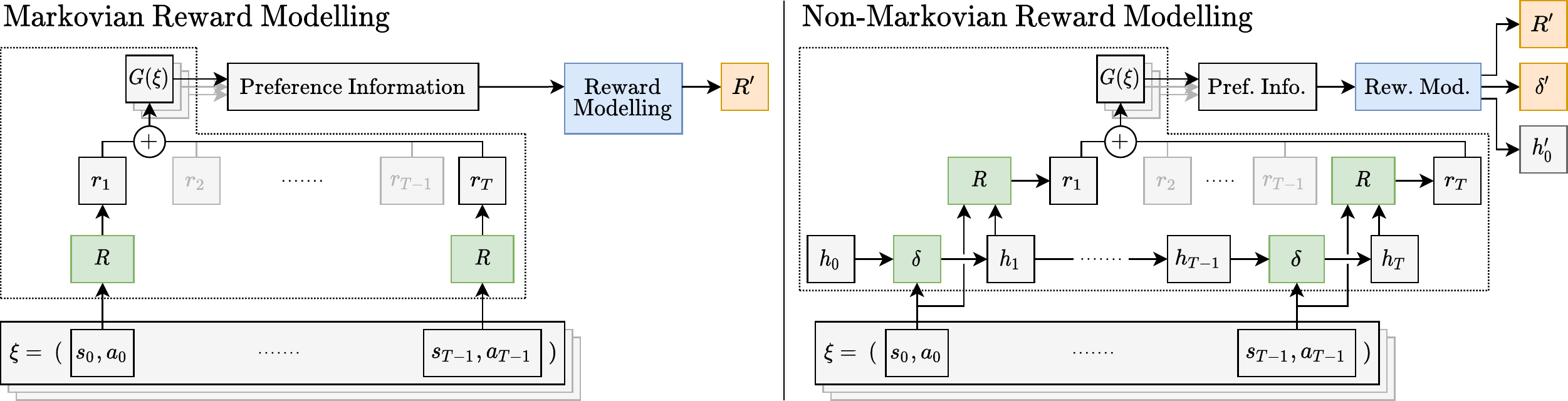}
    \caption{In Markovian RM, the human is assumed to sum $(+)$ over independent and equal reward assessments for the state-action pairs in a trajectory. In non-Markovian RM, per-timestep rewards additionally depend on hidden state information $h$ that accumulates over time.}
    \label{fig:problems}
\end{figure}

Consider an agent interacting with an environment with Markovian dynamics. At discrete time $t$, the current environment state $s_t\in\mathcal{S}$ and agent action $a_t\in\mathcal{A}$ condition the next environment state $s_{t+1}$ according to the dynamics function $D:\mathcal{S}\times\mathcal{A}\rightarrow\Delta\mathcal{S}$. A trajectory $\xi\in\Xi$ is a sequence of state-action pairs, $\xi=((s_0,a_0),...,(s_{T-1},a_{T-1}))$, and a human's preferences about agent behaviour respect a real-valued return function $G:\Xi\rightarrow\mathbb{R}$. In traditional (Markovian) RM, return is assumed to decompose into a sum of independent rewards over state-action pairs, $G(\xi)=\sum_{t=0}^{T-1}R(s_t,a_t)$, and the aim is to reconstruct $R'\approx R$ from possibly-noisy sources of preference information. In our generalised non-Markovian model, we consider the human to observe a trajectory sequentially and allow for the possibility of hidden state information that accumulates over time and parameterises $R$:
\begin{equation}
    G(\xi)=\sum_{t=0}^{T-1}R(s_t,a_t,h_{t+1})\ \ \ \text{where}\ \ \ h_{t+1}=\delta(h_t,s_t,a_t),
\end{equation}
$\delta$ is a hidden state dynamics function, and $h_0$ is a fixed value for the initial hidden state.
%Reward is no longer Markovian with respect to states and actions, and
Reconstruction of the human's preferences now requires the estimation of $\delta'\approx \delta$ and $h_0'\approx h_0$ alongside $R'\approx R$. We visualise the difference between Markovian and non-Markovian RM in Figure \ref{fig:problems}.

The hidden state $h$ may be interpreted as (1) an external feature of the environment that is detectable by the human but excluded from the state, or (2) a psychological feature of the person themselves, through which their response to each new observation is influenced by what they have seen already. The latter framing is more interesting for our purposes, and connects to the psychological literature on human judgement, memory, and biases \cite{kahneman2000evaluation}. In practice, hidden state information may encode the human's preferences about the order in which a sequence of behaviours should be performed, the effect of historic observations on their subjective mood (and in turn on their reward evaluations), or cognitive biases which corrupt the way they aggregate instantaneous rewards into trajectory-level feedback. All of these complications are liable to arise in practical RM applications, but cannot be handled when the Markovian reward assumption is made. Appendix \ref{app:use_cases} elaborates on this discussion, presenting motivating use cases and limitations of non-Markovian RM.

In this work, we focus on one of the simplest and most explicit forms of preference information: direct labelling of returns $G(\xi_i)$ for a dataset of $N$ trajectories $\{\xi_i\}_{i=1}^N$. We aim to solve the reconstruction problem by minimising the squared error in predicted returns:
\begin{equation} \label{eq:objective}
    \underset{R',\delta',h_0'}{\text{argmin}}\sum_{i=1}^N\left( G(\xi_i)-\sum_{t=0}^{T_i-1}R'(s_{i,t},a_{i,t},h_{i,t+1}')\right)^{\hspace{-0.2em}2}
    \text{where}\begin{array}{l}h_{i,0}'=h'_0\\h_{i,t+1}'=\delta'(h'_{i,t},s_{i,t},a_{i,t})\end{array}\forall i\in\{1..N\}.
\end{equation}

We observe that Equation \ref{eq:objective} perfectly matches the definition of a MIL problem. Each trajectory $\xi_i$ can be considered as an ordered bag of instances $((s_{i,0},a_{i,0}),...,(s_{i,{T_i}-1},a_{i,{T_i}-1}))$ with unobserved instance labels $R(s_{i,t},a_{i,t},h_{i,t+1})$, an observed bag label $G(\xi_i)=\sum_{t=0}^{T-1}R(s_{i,t},a_{i,t},h_{i,t+1})$, and temporal instance interactions via the changing hidden state $h_{i,t}$. This correspondence motivates us to review the space of existing MIL models (specifically those that model temporal dependencies among instances) to provide a starting point for developing our non-Markovian RM approach.

\vspace{-0.1cm}
\subsection{MIL RM Architectures}
\label{sec:models}

The MIL literature contains a variety of architectures for handling temporal instance dependencies, including graph neural networks (GNNs) \cite{tu2019multiple} and transformers \cite{shao2021transmil}. While effective for many problems, such architectures are an unnatural fit to non-Markovian RM as they contain no direct analogue of a hidden state $h'$ carried forward in time, instead handling dependencies via some variant of message-passing between instances. LSTM-based MIL architectures \citep{li2020multi, wang2020defense} provide a more promising starting point since they explicitly represent both $h'$ (implemented as a continuous-valued vector) and its temporal dynamics function $\delta'$ (a particular arrangement of gating functions). 

\begin{figure}[htb!]
    \centering
    \includegraphics[width=\textwidth]{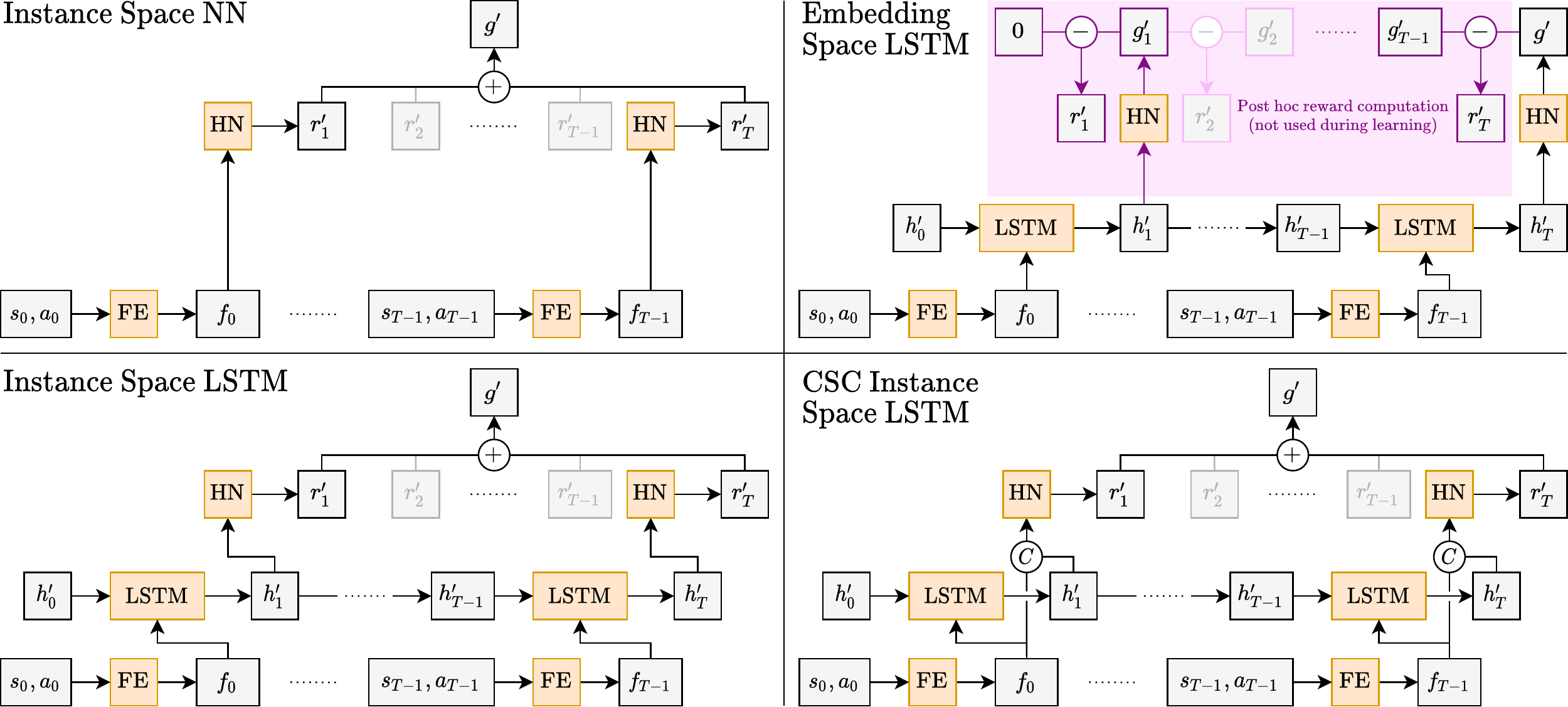}
    \caption{MIL architectures used in this work. FE = feature extractor; HN = head network; $(+)$ = scalar summation; $(-)$ = scalar subtraction; $(C)$ = vector concatenation.
    }
    \label{fig:architectures}
\end{figure}

\vspace{-0.1cm}
Starting from an existing LSTM-based MIL architecture, we propose two successive extensions as well as a na\"ive baseline that cannot handle temporal dependencies. All four architectures include a feature extractor (FE) for mapping state-action pairs into feature vectors and a head network (HN) that outputs predictions. These architectures are depicted in Figure \ref{fig:architectures}. Note we use the same nomenclature as \cite{carbonneau2018multiple} and \citep{wang2018revisiting}: \textit{embedding space} approaches produce an overall bag representation that is used for prediction, while \textit{instance space} approaches produce predictions for each instance in the bag and then aggregate those predictions to a final bag prediction.

\vspace{-0.2cm}
\paragraph{Base Case: Embedding Space LSTM} \hspace{-0.8em} This architecture, proposed by \citet{wang2020defense}, processes all instances in a bag sequentially and uses the final LSTM hidden state as a bag embedding. This is fed into the HN, which predicts the bag label $g'$ (return in the RM context). Although this model can account for temporal dependencies, it does not inherently produce instance predictions (rewards), which require some post hoc analysis to recover. While methods exist for computing instance importance values as a form of interpretability \cite{early2022model}, these are not guaranteed to sum to the bag label as stipulated by the reward-return formulation. We propose a new method: at time $t$, the predicted reward $r_t'$ is calculated by feeding the LSTM hidden state at times $t-1$ and $t$ into the HN to obtain two \textit{partial} bag labels/returns $g_{t-1}'$ and $g_{t}'$, and computing the difference of the two, i.e., $r_t' = g_t' - g_{t-1}'$. We define $g_0' = 0$. This post hoc computation is shown in purple in Figure \ref{fig:architectures}.

\vspace{-0.2cm}
\paragraph{Extension 1: Instance Space LSTM} \hspace{-0.8em} The post hoc computation of reward proposed above is rather inelegant and often yields poor predictions (see Section \ref{sec:results} and Appendix \ref{app:toy_datasets}), likely because rewards are never computed or back-propagated through during learning. This leads us to propose an improved architecture, which is structurally similar but differs in how network outputs are mapped onto RM concepts. The change places reward predictions on the back-propagation path. Given the LSTM hidden state at time $t$, the output of the HN is taken to be the instantaneous reward $r_t'$ rather than the partial return. Rewards are computed sequentially for all timesteps in a trajectory and summed to give the return prediction $g'$. We thereby obtain a model that both handles temporal dependencies and produces explicitly-learnt reward predictions.

\vspace{-0.2cm}
\paragraph{Extension 2: Concatenated Skip Connection (CSC) Instance Space LSTM} \hspace{-0.8em} In both of the preceding architectures, the LSTM hidden state $h'_t$ is the sole input to the HN. This requires $h'_t$ to represent all reward-relevant information from both the true hidden state $h_t$ and the latest state-action pair $s_{t-1},a_{t-1}$ to achieve good performance. To lighten the load on the LSTM, we further extend the Instance Space LSTM model with a skip connection \cite{he2016deep, huang2017densely} which concatenates the FE output onto the hidden state before feeding it to the HN. In principle, this should allow the hidden state to solely focus on representing temporal dependencies. As well as improving RM performance compared to an equivalent model without skip connections, we find in Section \ref{sec:hidden_embeddings} that this modification tends to yield more interpretable and disentangled hidden state representations.

\vspace{-0.2cm}
\paragraph{Markovian Baseline: Instance Space Neural Network (NN)} \hspace{-0.8em} To quantify the cost of ignoring temporal dependencies, we also run experiments with a baseline architecture that feeds only the FE output for each state-action pair into the HN, yielding fully-independent reward predictions which are summed to give the return prediction. This independent predict-and-sum architecture has precedence in both MIL, where it is referred to as mi-Net by \citet{wang2018revisiting}, and in RM, where it embodies the de facto standard Markovian reward assumption \cite{christiano2017deep}.

\subsection{Training Agents with Non-Markovian RM Models}
\label{sec:method_augmenting_rl}

\begin{figure}[htb!]
    \centering
    \includegraphics[width=\textwidth]{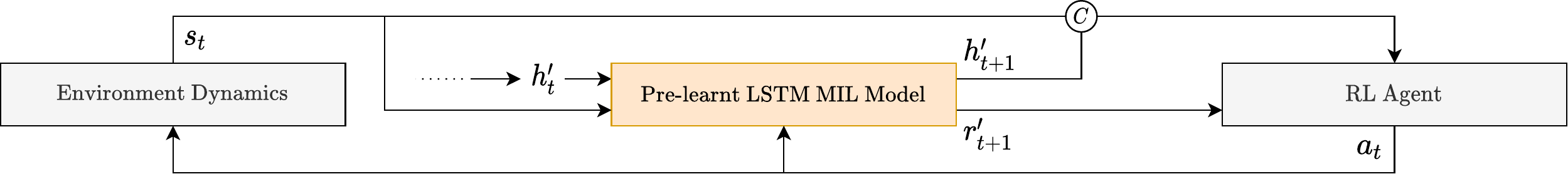}
    \caption{During RL agent training, our LSTM MIL models sit at the centre of the agent-environment loop by which states $s_t$ and actions $a_t$ are exchanged. We focus on episodic tasks, where the environment state periodically resets. The LSTM hidden state is simultaneously reset to $h'_0$ at the start of an episode, then is iteratively updated over time $t$ given the state-action pairs $s_t,a_t$. At time $t$ the environment state $s_t$ is augmented with the post-update hidden state $h'_{t+1}$ by concatenation, and this augmented state is observed by the agent. $s_t$, $a_t$ and $h'_{t+1}$ are used to compute a reward $r'_{t+1}$ following the relevant steps from Figure \ref{fig:architectures}, and the reward is also sent to the agent. In the language of NMRDPs, the hidden state augmentation \textit{expands} the agent's learning problem into an MDP by providing the additional information required to make the rewards Markovian. Note that unlike during learning of the MIL models, return predictions are never required. 
    % Also note the importance of a recurrent model architecture. As they lack a sequentially-updated hidden state, transformer- or GNN-based MIL models would be unsuitable for such online deployment.
    }
    \label{fig:training_setup}
\end{figure}

In this work, as in RM more widely, we are not solely interested in learning reward functions to represent human preferences, but also in the downstream application of rewards to train agents' action-selection policies. After optimising our LSTM-based models on offline trajectory datasets, we deploy them at the interface between conventional RL agents and their environments. Going beyond prior work, where a learnt model is used to either generate a reward signal for an agent to maximise \cite{christiano2017deep} or augment its observed state representation with hidden state information \cite{icarte2018using}, our models serve a dual role, providing \textit{both} rewards and state augmentations. Figure \ref{fig:training_setup} describes this setup in detail.

%\cite{icarte2018using,jarboui2021trajectory} as a guide for feeding the LSTM hidden state into RL architecture

%Non-Markovian reward functions (both oracles and learnt LSTM models) are implemented as wrappers on rewardless base environments. The role of a wrapper is to track the hidden state of either the oracle or the LSTM model over the course of an episode, and use this to compute rewards to return to the agent. In the ``no hidden state" case, we return the raw environment state $[x,y]$ to the agent unmodified. In the ``hidden state" case, we concatenate the hidden state onto the end of the environment state, thereby expanding the effective state space from the agent's perspective and making rewards Markovian. 

\vspace{-0.1cm}
\section{Experiments and Results}
\label{sec:results}

After initially validating our models on several toy datasets (see Appendix \ref{app:toy_datasets}), we focus the bulk of our evaluation on five RL tasks. As running experiments with people is costly, we use the standard RM approach of generating synthetic preference data (here trajectory return labels) using ground truth \textit{oracle} reward functions \cite{christiano2017deep} (for a discussion comparing the use of oracle and human labels, see Appendix \ref{app:human_vs_oracle_labelling}). Unlike prior work, these oracle reward functions depend on historical information that cannot be recovered from the instantaneous environmental state, thereby emulating the disparity between the information that a human evaluator possesses while viewing a trajectory sequentially, and that contained in the state alone. In this section, we introduce our RL tasks (Section \ref{sec:rl_datasets}), evaluate the quality of reward reconstruction (Section \ref{sec:results_reward_modelling}), investigate the use of MIL RM models for agent training (Section \ref{sec:results_rl_performance}), and evaluate their robustness to label noise (Section \ref{sec:robustness}).

%We focus our evaluation of the RM MIL models on a quantitative evaluation of their ability to predict return and reward, and a qualitative evaluation of their interpretability. For the former, we measure the mean squared error (MSE) between the predicted and true values, and for the latter, we analyse the hidden state embeddings learnt by the models. In the rest of this section, we first present our results using synthetic datasets (Section \ref{sec:synthetic_results}), and then RL datasets (Section \ref{sec:rl_results}).

\subsection{RL Task Descriptions}
\label{sec:rl_datasets}

\begin{figure}[htb!]
    \centering
    \includegraphics[width=\textwidth]{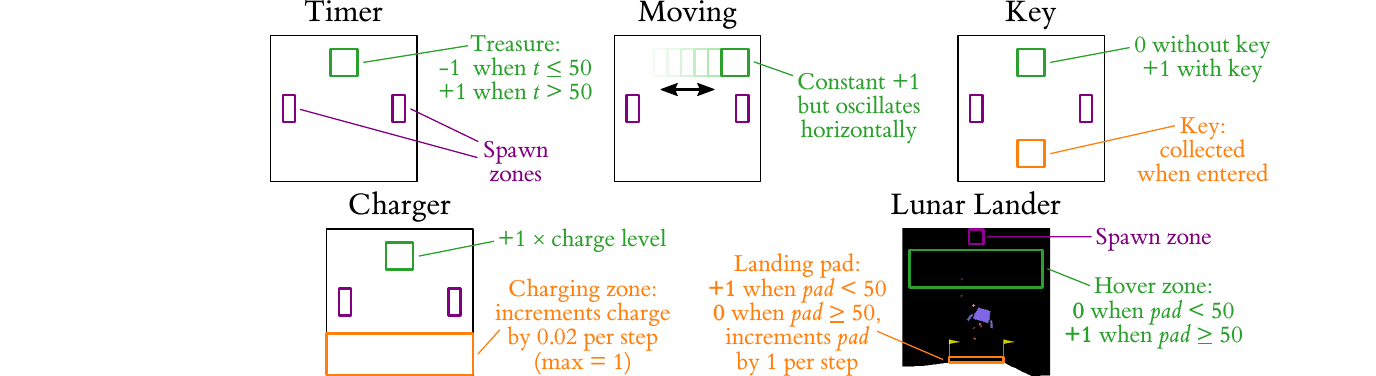}
    %\caption{Environment layouts for the four non-Markovian RL tasks.}
    \caption{Visualisations of the five non-Markovian RL tasks.}
    \label{fig:env_layouts}
\end{figure}

We apply our methods to five non-Markovian RL tasks, the first four of which are within a common 2D navigation environment and are specifically designed to capture different kinds of non-Markovian structure.
% To the best of our knowledge, no agreed-upon benchmarks for non-Markovian tasks exist (e.g., all standard OpenAI Gym \cite{gym} environments are Markovian), and prior work commonly focuses on grid worlds with very small discrete state spaces \citep{gaon2020reinforcement, littman2017environment}.
Each environment has two spawn zones
%(where the agent initialises)
and an episode time limit of $T=100$; see Figure \ref{fig:env_layouts}. In each case, the environment state contains the $x,y$ position of the agent only. The tasks involve moving into a \textit{treasure} zone, contingent on some hidden information that cannot be derived from the current $x,y$ position, but is instead a function of the full preceding trajectory. In the first two cases the hidden information varies with time only, but in the other two it depends on the agent's past positions.

\vspace{-0.3cm}
\paragraph{Timer} For times $t\leq 50$ the treasure gives a reward of $-1$ for each timestep that the agent spends inside it, before switching to $+1$ thereafter. Since time is not included in the environment state, recovering the reward function by only observing the agent's current position is impossible.

\vspace{-0.3cm}
\paragraph{Moving} The Timer task only captures a binary change, therefore we generalise it to be continuous. In this case, the treasure zone oscillates left and right at a constant speed. Again, this is not captured in the environment state, but can be recovered if the length of the preceding trajectory is known.

\vspace{-0.3cm}
\paragraph{Key} Before reaching the treasure zone, the agent must first enter a second zone to collect a key; otherwise it receives $0$ reward. As the key's status is not captured in the environment state, a temporal dependency exists between the agent's past positions and the reward it obtains from the treasure.

\vspace{-0.3cm}
\paragraph{Charger} We generalise the Key task by replacing the key zone with a charging zone that builds up the amount of reward the agent will receive when it reaches the treasure. The reward now depends not only on whether the agent visits a zone (binary), but how long it spends there (continuous).

For the fifth and most complex task, we adapt \textbf{Lunar Lander} from OpenAI Gym \cite{gym}, adding the condition that the lander should take off again and stably hover after $50$ timesteps on the landing pad. This is analogous to the Charger task but with a larger state-action space and longer episodes ($T=500$). Further details on the tasks and MIL model hyperparameters are given in Appendix \ref{app:tasks_data_models}.

An important design decision for the LSTM-based models is the size of the hidden state, as it affects both performance and interpretability. For all the above tasks, we know \textit{a priori} that it is possible to capture the temporal dependencies in at most two dimensions, so we constrain our models to use 2D hidden states. This allows us to visualise and interpret the hidden representations in Section \ref{sec:hidden_embeddings}. 

\subsection{Reward Modelling Results}
\label{sec:results_reward_modelling}

Below we discuss the performance of the reward reconstruction for the different MIL RM models on our five RL tasks. For each task, we generate initial trajectories to form our MIL RM datasets (see Appendix \ref{app:tasks_data_models}).
% 5000 trajectories of length 100 by deploying a random agent and post-screening to obtain a wide distribution of outcomes and return values (see Appendix \ref{app:tasks_data_models} for details on the screening process).
Results from MIL models trained on these trajectories are given in Table \ref{tab:rl_results}. We observe that the CSC Instance Space LSTM model is on average the best-performing model for predicting both trajectory returns and timestep rewards. While the Embedding Space LSTM model performs best at predicting return on the Key and Lunar Lander tasks (as is not constrained to the summation of reward predictions as in the other architectures), it struggles on the reward metric (due to the use of a proxy post hoc method). As it is important for these models to achieve strong performance on both return and reward prediction, the Instance Space LSTM and CSC Instance Space LSTM models are better candidates than the Embedding Space LSTM. Also note that the Instance Space NN that serves as our Markovian RM baseline performs very poorly on return prediction, indicating that these tasks indeed cannot be learnt without modelling temporal dependencies.

\setlength{\tabcolsep}{3pt}
\begin{table}[htb!]
	\centering
	\small
	\caption{MIL RM return (top) and reward (bottom) results, using ten repeats. The Lunar Lander results are the average test set MSE of the top five models (with scaling; see Appendices \ref{app:mil_ll} and \ref{app:mil_ll_discussion}). For the other tasks, each measurement is the test set MSE averaged over all ten repeats. The standard errors of the mean are given, and the Lunar Lander reward results are scaled by \num{1e-5}.}
    \label{tab:rl_results}
	\begin{tabular}{llllll}
		\toprule
		Model & Timer & Moving & Key & Charger & Lunar Lander \\
		\midrule
 		Instance Space NN & 130.8 $\pm$ 1.530 & 22.24 $\pm$ 0.441 & 7.764 $\pm$ 0.232 & 7.783 $\pm$ 0.214 & 2.297 $\pm$ 0.058 \\
		Embedding Space LSTM & 3.151 $\pm$ 0.662 & 13.04 $\pm$ 0.899 & \textbf{0.360 $\pm$ 0.055} & 0.689 $\pm$ 0.124 & \textbf{0.416 $\pm$ 0.048} \\
		Instance Space LSTM & 7.313 $\pm$ 2.627 & 11.13 $\pm$ 1.169 & 0.488 $\pm$ 0.062 & 0.628 $\pm$ 0.126 & 1.223 $\pm$ 0.431 \\
		CSC Instance Space LSTM & \textbf{0.605 $\pm$ 0.166} & \textbf{5.307 $\pm$ 0.299} & 0.391 $\pm$ 0.083 & \textbf{0.125 $\pm$ 0.012} & 0.501 $\pm$ 0.035 \\
		\greyrule
		Instance Space NN & 0.217 $\pm$ 0.001 & 0.068 $\pm$ 0.000 & 0.011 $\pm$ 0.000 & 0.025 $\pm$ 0.000 & 7.484 $\pm$ 0.861 \\
        Embedding Space LSTM & 101.8 $\pm$ 60.35 & 3.033 $\pm$ 0.715 & 0.010 $\pm$ 0.008 & 0.037 $\pm$ 0.016 & 120.2 $\pm$ 24.27 \\
        Instance Space LSTM & 0.263 $\pm$ 0.038 & 0.069 $\pm$ 0.005 & 0.002 $\pm$ 0.000 & 0.005 $\pm$ 0.001 & 9.336 $\pm$ 3.116 \\
        CSC Instance Space LSTM & \textbf{0.073 $\pm$ 0.016} & \textbf{0.026 $\pm$ 0.002} & \textbf{0.001 $\pm$ 0.000} & \textbf{0.001 $\pm$ 0.000} & \textbf{7.365 $\pm$ 1.032} \\
		\bottomrule
	\end{tabular}
	\vspace{-0.2cm}
\end{table}
\setlength{\tabcolsep}{6pt}

\subsection{RL Training Results}
\label{sec:results_rl_performance}

\begin{figure}[htb!]
    \centering
    \includegraphics[width=\textwidth]{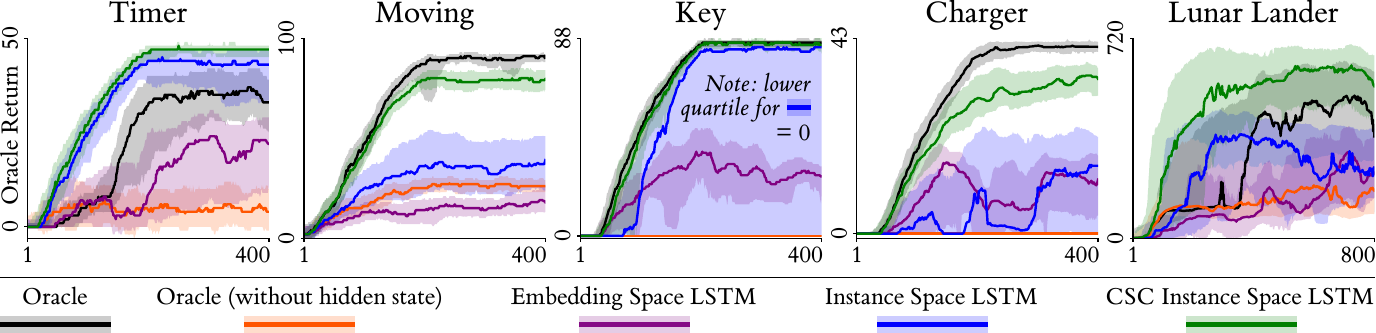}
    \caption{RL performance for different training configurations on our five RL tasks. The results given are the medians and interquartile ranges.
    % (across $20$-episode sliding windows)
    For the oracle results, we trained ten repeats, and for the MIL-LSTM results, we performed one RL training run for each MIL-LSTM model repeat.}
    \label{fig:rl_training_results}
\end{figure}

Following the method in Section \ref{sec:method_augmenting_rl}, we then train Soft Actor-Critic \cite{haarnoja2018soft} (Lunar Lander) and Deep Q-Network \citep{mnih2015human} (all others) RL agents to optimise the rewards learnt by the LSTM-based models. We evaluate agent performance in a post hoc manner by passing its trajectories to the relevant oracle. This evaluation provides an end-to-end measure of both reward reconstruction and policy learning, and is standard in RM \cite{christiano2017deep}. We baseline against agents trained with access to: a) the oracle reward function and the oracle hidden states, and b) just the oracle reward function without hidden states (i.e, using only the environment states that are missing information). In Figure \ref{fig:rl_training_results}, we observe that the CSC Instance Space LSTM model enables the best RL agent performance, coming closest to the oracle. Interestingly, for the Timer and Lunar Lander tasks, the CSC Instance Space LSTM model actually outperforms the use of the oracle, suggesting that the learnt hidden states are easier to exploit for policy learning than the raw oracle state (we investigate what these models have learnt in Section \ref{sec:hidden_embeddings}). 
% Furthermore, we observe a strong correlation between the performance of quality of the reward construction (Table \ref{tab:rl_results}) and performance of the corresponding RL agents, suggesting our MIL RM approach is effective for augmenting RL agent training. 
Note the poor performance of agents trained without hidden state information, which aligns with expectations. For further details on agent training, see Appendix \ref{app:rl_training_appendix}.

\vspace{-0.1cm}

\begin{figure}[htb!]
    \centering
    \includegraphics[width=\textwidth]{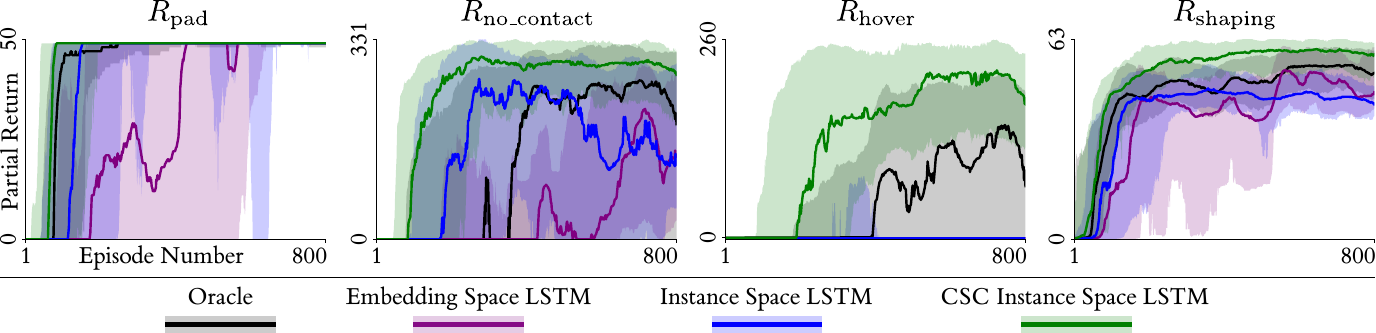}
    \caption{Decomposed oracle return curves for Lunar Lander.
    }
    \label{fig:decomposed_return_lunar_lander}
\end{figure}

\vspace{-0.1cm}
For Lunar Lander, we perform a deeper analysis of RL training performance, by decomposing the oracle return curves from Figure \ref{fig:rl_training_results} into the four reward components $R_\text{pad}$, $R_\text{no\_contact}$, $R_\text{hover}$ and $R_\text{shaping}$ (see Appendix \ref{app:rl_tasks} for definitions). The decomposed curves, shown in Figure \ref{fig:decomposed_return_lunar_lander}, allow us to diagnose the origins of the performance disparity between runs using different LSTM model architectures. There is relatively little separation in performance on the shaping reward $R_\text{shaping}$ and pad contact reward $R_\text{pad}$ (for the latter, all runs end up reliably achieving the maximum possible reward of $49$, although those using Embedding Space LSTM models require significantly more training time). This suggests that all models have been able to recover these components with reasonable fidelity. However, there are marked differences in performance on $R_\text{no\_contact}$ and $R_\text{hover}$ (the components relating to the second task stage of taking off and moving to the hover zone). For $R_\text{hover}$, runs using the CSC Instance Space LSTM peak at a return of around $200$ from this component, while those using the other two models almost never achieve non-zero return, i.e., only the RL agents trained using the CSC Instance Space LSTM RM models reliably learn to hover. This indicates that the models have learnt very different representations of reward and hidden state dynamics, which are effective for policy learning in the case of the CSC model, and highly ineffective for the others. 

Observe that runs using CSC Instance Space LSTM models outperform those with direct access to the ground truth oracle on all components, and most markedly on $R_\text{hover}$. This counterintuitive finding suggests that this model reliably learns hidden state representations that are easier for RL agents to leverage for policy learning than the ground truth ones, and potentially that certain errors in the reward prediction may actually be beneficial for the purpose of helping agents to complete the underlying task (especially the hovering stage). In typical RL parlance: the model's reward function appears to be better \textit{shaped} than the ground truth. The potential origins of this better-than-oracle phenomenon are investigated in Figure \ref{fig:lunar_lander_interpretability} (Appendix \ref{app:other_task_interpretability}).

\vspace{-0.1cm}

% \subsection{Hidden State Recovery Analysis}

% Hidden state dimensionality reduction and clustering. Note that \cite{jarboui2021trajectory} have similarly plotted hidden states to assess separation of different tasks.

% See the series of papers Learning/Understanding/Re-understanding Finite-State Representations of Recurrent Policy Networks \cite{koul2019learning}

% ---------------------------------------------------------------
%   ROBUSTNESS STUDY - COMMENT OUT IF MOVING TO APPENDIX
% ---------------------------------------------------------------
\vspace{-0.1cm}
\subsection{Robustness to Mislabelling}
\label{sec:robustness}

\begin{figure}[h!]
    \centering
    \includegraphics[width=\textwidth]{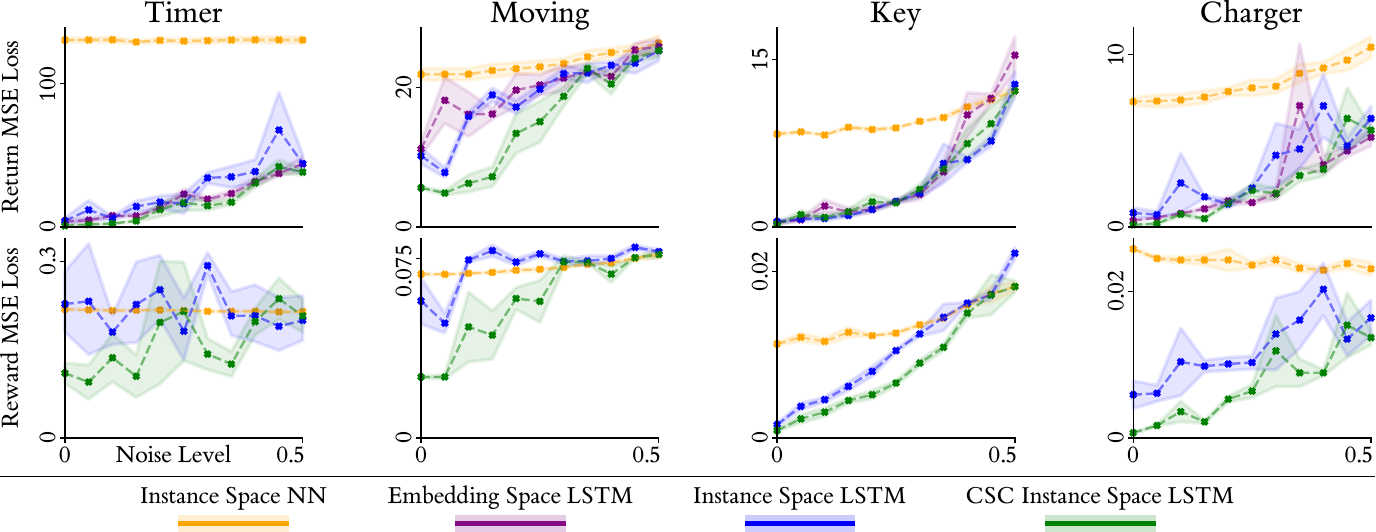}
    \caption{Performance of MIL RM models subject to label noise. We omit the Embedding Space LSTM reward losses as they are very high, and the Lunar Lander task due to long training times.}
    \label{fig:robustness_study}
\end{figure}

In this work, the return labels are provided by oracles rather than real people. When using human evaluators, there is likely to be uncertainty in the labels, and it is important to evaluate model robustness against such noise \cite{lee2021bpref}. We implement noise through label swapping \citep{rolnick2017deep}; this ensures the marginal label distribution remains the same and does not include out-of-distribution returns. In Figure \ref{fig:robustness_study}, we show how both return and reward prediction decay with noise levels increasing from 0 (no labels swapped) to 0.5 (half swapped). The rate, smoothness, and consistency (across three repeats) of this decay varies between tasks, with decays in return prediction generally being smoother. We observe that the CSC Instance Space LSTM model remains the strongest predictor of both return and reward in the majority of cases, indicating general robustness and providing evidence that the model should still be effective with imperfect human labels. On all metrics aside from Timer reward loss (where the mix of negative and positive rewards makes the effect of noise especially unpredictable), a noise level of at least 0.3 is required for the CSC Instance Space LSTM model to perform as badly as the Instance Space NN baseline does with no noise at all.
%Also, the Timer reward prediction is sensitive to noise as this task involves both negative and positive rewards (all the other tasks produce reward $\geq 0$), meaning the label noise is particularly destructive. 

\vspace{-0.1cm}

% ---------------------------------------------------------------
%   ROBUSTNESS STUDY - COMMENT OUT IF MOVING TO APPENDIX
% ---------------------------------------------------------------

\vspace{-0.2cm}
\section{Discussion}
\label{sec:discussion}

In this section, we seek to interpret our MIL RM models, analysing the distribution of learnt hidden states (Section \ref{sec:hidden_embeddings}) as well as their temporal dynamics over the course of a trajectory (Section \ref{sec:probing}). Finally, in Section \ref{sec:future_work} we discuss the limitations of this work and potential areas for future work.

\vspace{-0.1cm}
\subsection{Hidden State Analysis}
\label{sec:hidden_embeddings}
\vspace{-0.1cm}

\begin{figure}[htb!]
    \centering
    \includegraphics[width=\textwidth]{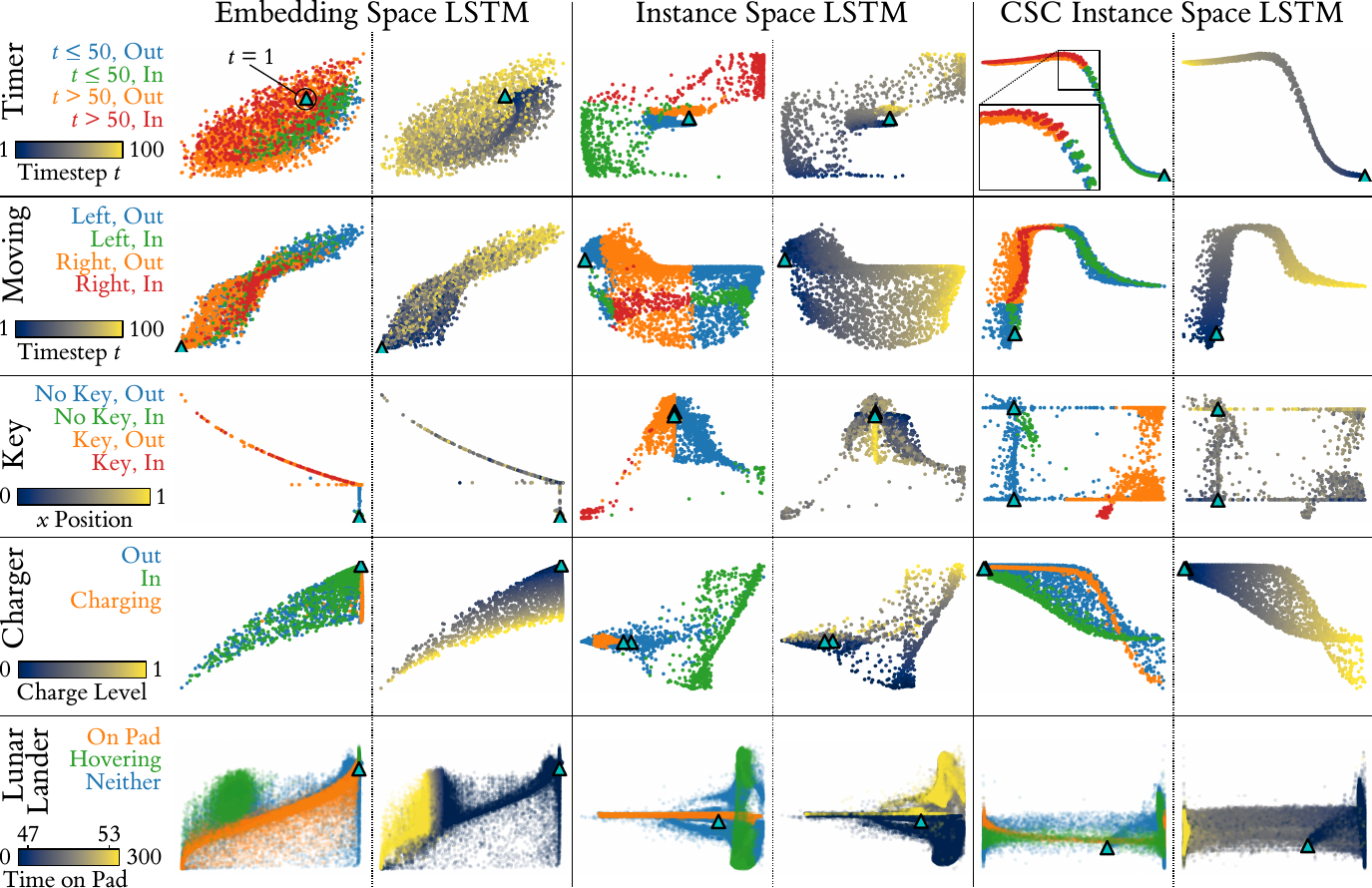}
    \vspace{-0.5cm}
    \caption{Learnt hidden state embeddings for our MIL RM models. For each model and task, we categorise the hidden state embeddings depending on the true environment state (first column for each model). In and Out environments states indicate whether the agent is in the treasure zone or not, and for the Moving task, Left and Right indicate the direction in which the goal is currently moving. We also provide labelling based on temporal information (second column for each model). Furthermore, we include markers to indicate the hidden states for the centres of the agent spawn zones. In each case, we elected to use the best-performing repeat for each model as assessed by the reward reconstruction (see Table \ref{tab:rl_results}). Note, for the Key task, as the temporal information is captured in the state categorisation (No Key vs Key), we use the second column to show the relationship between the hidden embeddings and the agent's $x$ position.
    % This helps showcase the separation of the two possible start spaces learnt by the CSC Instance Space LSTM model.
    }
    \label{fig:hidden_state_comparison}
\end{figure}

The primary purpose of RM is to perform accurate reward reconstruction to facilitate agent training, but there is a secondary opportunity to improve understanding of human preferences through interpretability analysis of the learnt models. We can directly visualise the 2D LSTM hidden states of our oracle experiments, which enables a qualitative comparison of the various model architectures (see Figure \ref{fig:hidden_state_comparison}). Visualising the hidden states with respect to the temporal dependencies indicates that the CSC Instance Space LSTM model has learnt insightful hidden state representations. Breaking down the CSC Instance Space LSTM model hidden embeddings: for the Timer task, time is represented along a curve, with a sparser representation around $t=50$ (the crossover point when the treasure becomes positive). For the Moving task, time is similarly captured along with an additional notion of the change in treasure direction from right to left. For the Key task, the binary state of no key vs key is separated, with additional partitioning based on $x$ position, denoting the two different start points of the agent. In the Lunar Lander task, the model has learnt a strong separation between states either side of the crossover point when the time on the pad is equal to $50$, with high sparsity around the crossover point. In comparison, the Embedding and Instance Space LSTM models have not learnt as sparse a representation. We discuss the Charger task in Section \ref{sec:probing}.

% \N{This is due to the model's skip connections \textemdash{} by concatenating the state instance embedding to the hidden embedding prior to instance prediction classification, the hidden embedding can focus on solely capturing the temporal information. However, we note that the CSC Instance Space LSTM hidden embeddings still contain some information pertaining to agent position (demonstrating by colouring the points according to the true environment state; left columns in Figure \ref{fig:hidden_state_comparison}), suggesting encoding solely the temporal information may not be the optimal approach.}

% \N{better representations for the hidden state embeddings, i.e, the temporal dependencies are clearer} and the different hidden state \N{categories} are better separated, enabling the RL agent to easily access the additional information it requires to achieve a high level of performance. 

% This is due to the concatenated skip connections \textemdash{} by including the instance embedding in the instance classification, the hidden embedding only has to capture the temporal information.

% \begin{figure}[htb!]
%     \centering
%     \includegraphics[width=\textwidth]{figures/interpretability_comparison_combined.png}
%     \caption{Hidden state comparison. \N{Make text bigger and remove `treasure' from dataset names.}}
%     \label{fig:hidden_state_comparison_old}
% \end{figure}

\vspace{-0.3cm}
\subsection{Trajectory Probing}
\label{sec:probing}
\vspace{-0.1cm}

\begin{figure}[htb!]
    \centering
    \includegraphics[width=\textwidth]{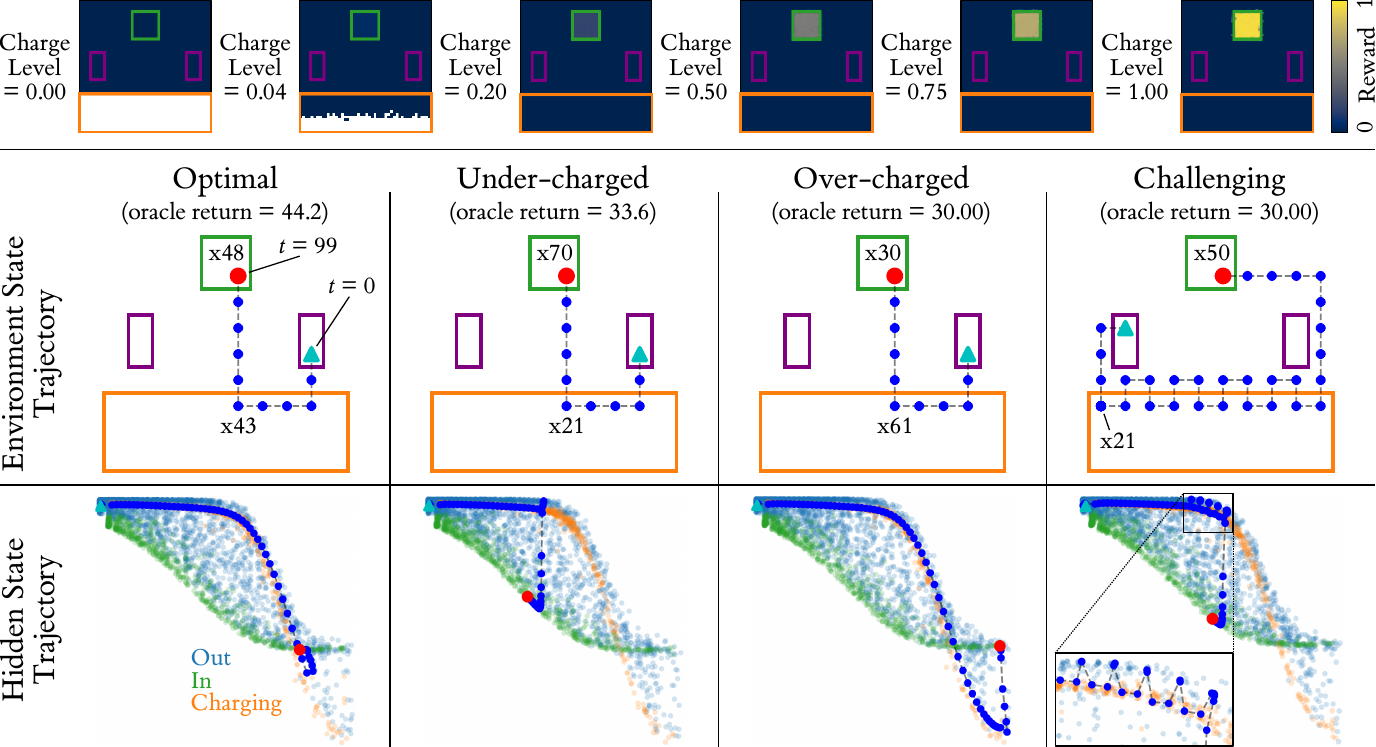}
    \vspace{-0.4cm}
    \caption{Interpretability for the CSC Instance Space LSTM model on the Charger task. \textbf{Top}: the learnt relationship between agent position, charge level, and reward. \textbf{Middle/bottom}: Four probe trajectories demonstrating hidden state transitions as the trajectory progresses. \textit{Optimal}: best possible return (charge to sufficient level then maximise time in treasure). \textit{Over-charged}: continuing to charge after maximum charge of $1$ is reached. ``\texttt{x}$n$'' labels indicate the agent remains in a position for $n$ timesteps. As in Figure \ref{fig:hidden_state_comparison}, we use the best-performing model according to reward reconstruction.}
    \label{fig:charger_interpretability}
\end{figure}

We further interpret our models by visualising the learnt reward with respect to the environment state, and by using hand-specified \textit{probe} trajectories to verify that the learnt hidden state transitions mimic the true transitions. We present the above for the CSC Instance Space LSTM model on the Charger task in Figure \ref{fig:charger_interpretability} (Appendix \ref{app:other_task_interpretability} contains similar figures for all other tasks). The top row shows that the model has correctly learnt the relationships between position, charge, and reward (reward increases in the treasure zone as charge increases). From the probes, we can see how the charge level can be recovered from the hidden states. We also note that the inflection point between under-charging and over-charging is captured, i.e., this is where the optimal charge level lies, subject to some noise based on where the agent starts in the spawn zones. Furthermore, with the Challenging probe, we observe that the learnt hidden states align with the agent moving in and out of the charging zone.

\vspace{-0.05cm}

% \begin{figure}[htb!]
%     \centering
%     \includegraphics[width=\textwidth]{figures/charger_interpretability.pdf}
%     \caption{Interpretability for best-performing CSC on Charger}
%     \label{fig:charger_interpretability}
% \end{figure}

\vspace{-0.2cm}
\subsection{Limitations and Future Work}
\label{sec:future_work}

Although we analyse the performance of our methods in the presence of noisy labels in Section \ref{sec:robustness}, a major area of future work is to apply our methods to human labelling (for a discussion of this, see Appendix \ref{app:human_vs_oracle_labelling}).
% potentially as a diagnostic tool to understand the extent to which real preferences are non-Markovian, and to generalise the learning
% which would require generalisation 
% to other information sources such as trajectory demonstrations \citep{ng2000algorithms} and pairwise choices \cite{christiano2017deep}.
Another area of future work involves more complex environments, for example the use of tasks with image observations, similar to the Atari environments in Open AI Gym \citep{gym}.
% We have proposed benchmark non-Markovian tasks, but additional experiments on more complex environments would be insightful \textemdash{} one approach would be to adapt other established OpenAI Gym benchmarks \citep{gym} to be non-Markovian.
Furthermore, we perform RM from either an offline dataset or from only one RL training iteration; an iterative bootstrapping approach with multiple RL + RM training iterations could lead to improved RL results. There are also limitations with our MIL RM approach for the Lunar Lander task; see Appendix \ref{app:mil_ll_discussion} for details and suggestions for future work. More generally, we hope that our identification of the link between RM and MIL may inspire a bidirectional transfer of tools and techniques.

% Requirement for rebalanced pilot data: for practical applications may want to iteratively bootstrap, as done in prior work on interactive RL. Also, since we have added complexity to the reconstruction problem we have used the simplest possible feedback mechanism: direct return labels. 
% Our model uses a constant, albeit learnable, hidden state initialisation at the start of an episode, if psychological framing, may be persistent from previous queries

% Ours of one of many possible architectures that mirror existing computational models of humans' departure from unweighted summation during temporal reward aggregation, e.g., peak-end \citep{kahneman2000evaluation}, leaky integration \cite{vestergaard2015choice}. Compare architectures on real human data as a diagnostic tool: explore the extent to which preferences are non-Markovian.

\vspace{-0.2cm}
\section{Conclusion}
\label{sec:conclusion}

\vspace{-0.05cm}
We posed the problem of non-Markovian RM, which removes an unrealistic assumption about how humans evaluate temporally extended agent behaviours. After identifying an isomorphism between RM and MIL, we proposed and evaluated novel MIL-inspired models that allow us to reconstruct non-Markovian reward functions, augment agent training, and interpret their learnt representations.

\vspace{-0.2cm}
\section*{Acknowledgements}

\vspace{-0.05cm}
This work was funded by the AXA Research Fund, the UKRI Trustworthy Autonomous Systems Hub (EP/V00784X/1), and an EPSRC/Thales industrial CASE award. We would also like to thank the Universities of Southampton and Bristol, as well as the Alan Turing Institute, for their support. The authors acknowledge the use of the IRIDIS (Southampton) and BlueCrystal (Bristol) high-performance computing facilities and associated support services in the completion of this work.

\newpage
\bibliographystyle{abbrvnat}
{\fontsize{9}{0}\bibliography{bibliography}}

%%%%%%%%%%%%%%%%%%%%%%%%%%%%%%%%%%%%%%%%%%%%%%%%%%%%%%%%%%%%
\section*{Checklist}

\begin{enumerate}

\item For all authors...
\begin{enumerate}
  \item Do the main claims made in the abstract and introduction accurately reflect the paper's contributions and scope?
    \answerYes{}
  \item Did you describe the limitations of your work?
    \answerYes{See Section \ref{sec:future_work}}
  \item Did you discuss any potential negative societal impacts of your work?
    \answerNA{}
  \item Have you read the ethics review guidelines and ensured that your paper conforms to them?
    \answerYes{}
\end{enumerate}

\item If you are including theoretical results...
\begin{enumerate}
  \item Did you state the full set of assumptions of all theoretical results?
    \answerNA{}
   \item Did you include complete proofs of all theoretical results?
    \answerNA{}
\end{enumerate}

\item If you ran experiments...
\begin{enumerate}
  \item Did you include the code, data, and instructions needed to reproduce the main experimental results (either in the supplemental material or as a URL)?
    \answerYes{See Appendix \ref{app:reproducibility}}
  \item Did you specify all the training details (e.g., data splits, hyperparameters, how they were chosen)?
    \answerYes{See Appendices \ref{app:toy_datasets}, \ref{app:tasks_data_models}, and \ref{app:rl_training_appendix}}
    \item Did you report error bars (e.g., with respect to the random seed after running experiments multiple times)?
    \answerYes{See Appendix \ref{app:reproducibility}}
        \item Did you include the total amount of compute and the type of resources used (e.g., type of GPUs, internal cluster, or cloud provider)?
    \answerYes{See Appendix \ref{app:reproducibility}}
\end{enumerate}

\item If you are using existing assets (e.g., code, data, models) or curating/releasing new assets...
\begin{enumerate}
  \item If your work uses existing assets, did you cite the creators?
    \answerNA{}
  \item Did you mention the license of the assets?
    \answerNA{}
  \item Did you include any new assets either in the supplemental material or as a URL?
    \answerYes{Our source code and models, which are included in the supplementary material and will be made available publicly upon acceptance.}
  \item Did you discuss whether and how consent was obtained from people whose data you're using/curating?
    \answerNA{}
  \item Did you discuss whether the data you are using/curating contains personally identifiable information or offensive content?
    \answerNA{}
\end{enumerate}

\item If you used crowdsourcing or conducted research with human subjects...
\begin{enumerate}
  \item Did you include the full text of instructions given to participants and screenshots, if applicable?
    \answerNA{}
  \item Did you describe any potential participant risks, with links to Institutional Review Board (IRB) approvals, if applicable?
    \answerNA{}
  \item Did you include the estimated hourly wage paid to participants and the total amount spent on participant compensation?
    \answerNA{}
\end{enumerate}

\end{enumerate}

%%%%%%%%%%%%%%%%%%%%%%%%%%%%%%%%%%%%%%%%%%%%%%%%%%%%%%%%%%%%

\newpage
\appendix

\setcounter{table}{0}
\renewcommand{\thetable}{A\arabic{table}}
\renewcommand{\theHtable}{A\arabic{figure}} % Needed for hyperref to work

\setcounter{figure}{0}
\renewcommand{\thefigure}{A\arabic{figure}}
\renewcommand{\theHfigure}{A\arabic{figure}} % Needed for hyperref to work

\section{Implementation and Resource Details}
\label{app:reproducibility}

This work was implemented in Python 3.8 / 3.10 and the machine learning functionality used PyTorch. All required libraries for our work are given in a \texttt{requirements.txt} file. Our code is publicly accessible at \url{https://github.com/JAEarly/MIL-for-Non-Markovian-Reward-Modelling}. The majority of MIL model training was carried out on a remote GPU service using a Volta V100 Enterprise Compute GPU with 16GB of VRAM, which utilised CUDA v11.0 to enable GPU support  (IRIDIS 5, University of Southampton). For the Lunar Lander task, training each MIL model took a maximum of eight hours. For the other tasks, this was a maximum of two hours. Trained models are included alongside the code. Fixed seeds were used to ensure consistency of dataset splits between training and testing; these are included in the scripts that are used to run the experiments. All our datasets were generated from code; both the scripts to generate the data and also the derived datasets themselves are included alongside our model training code. Dataset generation, as well as all RL agent training, was conducted on a second remote GPU service using a compute node with two Nvidia Pascal P100 cards. Data generation took a maximum of three hours per dataset (BlueCrystal Phase 4, University of Bristol). Agent training for the 2D navigation tasks was computationally light, requiring 8-12 minutes per 400-episode run, although we completed ten repeat runs for each permutation of task and MIL model architecture (one per MIL training repeat). 800-episode RL runs for Lunar Lander took approximately two hours each. Further details on executing the scripts to reproduce our results can be found in the \texttt{README.md} file in our code submission.

\section{Use Cases for Non-Markovian Reward Modelling}
\label{app:use_cases}

%We can interpret the hidden information as an external feature of the environment that is visible to the human but excluded from the state, or as a psychological state of the person themselves, through which their response to a new observation is influenced by what they have seen already. Appendix B elaborates on this discussion, presenting motivating use cases and limitations of non-Markovian RM.

The non-Markovian reward formulation applies to cases where rewards depend on hidden state information $h_t$ in addition to environment states $s_t$ and actions $a_t$, and this information is a function of previous state-action pairs but \textit{not} vice versa (i.e., there is no causal path from $h_t$ to $s_{t+k}$ for any $k>1$). This crucial caveat distinguishes the formulation from the more general class of partially observable MDPs and demarcates the set of domains to which it can be applied: those involving a secondary Markovian system that ``spectates'' on events in the environment without intervening. In the RM context, this secondary system is a black box (making its internal state $h_t$ hidden) and explicit rewards are unavailable, being replaced by a sparser and potentially noisy form of reward-dependent feedback (trajectory return labels in our work). Below we identify three classes of use case which fit this technical specification and provide one concrete example for each:

\paragraph{Ambiguous Subtasks} Cases involving an extended task with a sequential structure, where it is hard to formally define the conditions for subtask completion, but RM is feasible because a human ``knows it when they see it''. Here, the hidden state to be learnt represents the current subtask and any auxiliary information needed to determine its completion status.
\begin{itemize}
    \item \textbf{Concrete Example:} Using judges' scores to learn a performative display (e.g., gymnastics, aerobatics) chaining several manoeuvres whose start and end conditions are difficult to formalise \textit{a priori}. This could be considered as an extension of the single backflip task studied in the foundational RM work by \citet{christiano2017deep}.
\end{itemize}
\paragraph{Dependencies on Subjective Affect} Cases where a human's reward function is dependent on their affective (emotional) status, which in turn depends on their prior experiences. Assuming this information is not directly available in the observed environment state, it must be inferred from data.
\begin{itemize}
    \item \textbf{Concrete Example:} Using periodic satisfaction ratings to train a personal assistant robot whose owner's mood, needs and preferences vary from day to day. These variations may influence the preferred driving style of a chauffeur service or choice of evening meal.
\end{itemize}

\paragraph{Irrationalities/Cognitive Biases} Cases where one or more forms of bias colour a human's post hoc rating of an observed trajectory, even if their instantaneously-experienced reward is Markovian. Psychological studies of how humans aggregate immediate rewards into retrospective evaluations of the quality of an experience find that a straight summation assumption is unrealistic, with subjects exhibiting high sensitivity to contrast effects and recency bias (collectively termed the peak-end rule) \citep{kahneman2000evaluation}, and factoring in anticipated future states in addition to those actually observed \citep{ariely2000gestalt}. 
%illusion of control, optimism/pessimism, prospect bias and hyperbolic discounting
Here, the hidden state captures an aggregate representation of the biases at play in a given human's evaluation.\footnote{A fascinating philosophical question arises here. When (following the method of Section \ref{sec:method_augmenting_rl}) an RL agent is trained using a non-Markovian RM model that captures a cognitive bias, should the agent learn to maximise rewards \textit{including} the bias (which, for example, might lead it to prioritise its peak and final reward rather than seek uniformly good performance), or \textit{exluding} it (which would revert to uniform prioritisation). This issue of whether intelligent agents should seek to exploit human irrationalities when optimising for their revealed preferences, or appeal to their unbiased ``better angels'', relates to distinctions between first- and second-order preferences \cite{stanovich2008higher} or between experienced and remembered (or decision) utility \cite{berridge2014experienced}. We defer this question to those with more relevant expertise but note that regardless of the answer, it is essential to include the biases in the reward model using a method such as the one proposed in this work.}
\begin{itemize}
    \item \textbf{Concrete Example:} Using customer ``star ratings'' to improve a holiday planning agent whose recommendations aim to account for an unknown mix of biases such as the peak-end rule. The agent may use the learnt bias model to prioritise key moments in a holiday when managing the travel schedule and budget, in order to maximise future customers' star ratings.
\end{itemize}

Finally, we note that it is a matter of taste as to whether hidden state information is framed as situated \textit{inside a human evaluator's mind} or \textit{in the environment but only visible to the human}. It is not technically necessary to decide between these two framings, as the mathematical problem of non-Markovian RM is equivalent. From an agent's perspective, a human evaluator is part of an augmented environment, even if they never intervene directly to influence the state.

\section{Comparing Human and Oracle Labelling}
\label{app:human_vs_oracle_labelling}

Oracle-based experiments are often used to evaluate RM methods since they enable scalable quantitative validation \citep{griffith2013policy, hadfield2017inverse, reddy2020learning}. However, we identify three concrete differences between our oracle preference labelling method and realistic human labelling: 1) preference form, 2) preference sparsity, and 3) preference noise. Preference form is the different ways of providing labels; in this work we used return values which are highly informative and easy to learn from, but it has long been understood that humans find it easier to give less direct feedback, such as pairwise rankings \cite{christiano2017deep} or good/bad/neural labels \citep{kendall1948rank}. Preference sparsity occurs when the time- and cost-expensiveness of eliciting human labels reduces the proportion of data that can be labelled, and preference noise arises from uncertainties in the human labelling process (as opposed to perfect oracle labelling). We decided to focus on noise in this work as it is an established way of making oracle experiments more realistic \citep{lee2021bpref}, and also fits in with our discussion of human uncertainties and cognitive biases. Our experiments in Section \ref{sec:robustness} indicate that our methods degrade gracefully in the presence of noise, which gives us some confidence that they will transfer well to human labels. However, future work should consider preference sparsity and form, the latter of which will involve modifying the data collection pipeline and loss function (e.g., to a contrastive loss in the case of pairwise rankings). Beyond accounting for these three differences whilst still using oracle labels, the next step would be to conduct evaluations using actual human labels.

\section{Preliminary Experiments on Toy Datasets}
\label{app:toy_datasets}

In this section, we give detail our preliminary experiments that were run on toy problems to initially develop and validate our approach. Below we outline the datasets (Section \ref{sec:app_toy_datasets}), models and training hyperparameters (Section \ref{sec:app_toy_models}), and results (Section \ref{sec:app_toy_results}) of these experiments.

\subsection{Datasets}
\label{sec:app_toy_datasets}

We introduce three toy datasets, each abstracted from the RL context, to act as benchmark tests for our models. Each of these datasets uses ordered bags comprised of two-dimensional instances, where each instance has an associated label (called ``reward'' below, for consistency), and the overall bag label (return) is the sum of the instance labels.

\paragraph{Toggle Switch} An instance is the position of a toggle switch $ts$ and a value $v$; if the switch is on ($ts=1$), then the reward for the instance is $v$; otherwise the reward for the instance is $0$. Here, there is no hidden information, as it is possible to calculate the reward for an instance from its contents $ts,v$ alone. This serves as a null example to elucidate what happens in a Markovian setting.

\vspace{-0.2cm}
\paragraph{Push Switch} We modify the toggle switch setup so that the instance now represents a push switch ($p=1$ if this switch is pressed), where pressing the switch flips a binary hidden state $ts$ (the same information as was previously represented by the toggle switch). This hidden state then determines the reward as before ($v$ if $ts=0$, else $0$). As $ts$ must be tracked between successive instances, it is not possible to determine the reward for an instance solely by observing its contents $p,v$, so this setup is non-Markovian.

\vspace{-0.2cm}
\paragraph{Dial} We generalise the hidden state from a binary switch to a continuous-valued dial. Given an instance $m,v$, the dial's current value $d$ is moved up or down by $m$. The reward is then given as $d\cdot v$ The problem remains non-Markovian, but now the hidden state that needs tracking, $d$, is continuous rather than discrete.

\vspace{0.5cm}
% A summary of these three synthetic datasets is given in Table \ref{tab:synthetic_datasets}.

% \begin{table}[htb!]
%   \caption{Synthetic Oracle Datasets. \N{Finish caption}. \N{Ensure notation and naming is consistent with the respect of the paper.}}
%   \label{tab:synthetic_datasets}
%   \centering
%   \small
%   \begin{tabular}{lllll}
%     \toprule
%     Name & Input & Initial State & State Update & Reward \\
%     \midrule
%     Toggle Switch & $t$ (0 or 1); $v$ [0, 1] & None & None & $r = t * v$ \\
%     Push Switch & $p$ (0 or 1); $v$ [0, 1] & $t = 0$ & $t = 1-p$ & $r = t * v$ \\
%     Dial & $m$ [-0.5, 0.5]; $v$ [-0.5, 0.5] & $d = 0$ & $d = d+m$ & $r = d + v$ \\
%     \bottomrule
%   \end{tabular}
% \end{table}

\vspace{-0.4cm}
\subsection{Models and Hyperparameters}
\label{sec:app_toy_models}

When training the MIL models on the toy dataset, we used the Adam optimiser with a batch size of one (i.e., one bag per batch) to minimise mean squared error (MSE) loss. Training was performed using validation loss early stopping, i.e., if the validation loss did not decrease after a certain number of training epochs (patience value), we terminated the training and selected the model at which the validation loss was lowest. If the patience value was not reached (i.e., the validation loss kept decreasing), we terminated training after a maximum number of epochs had been reached, and again selected the model at which the validation loss was lowest. The hyperparameters for training the models on each dataset (including learning rate (LR) and weight decay (WD)) are given in Table \ref{tab:toy_train_hyperparams}. Dropout was not used. These hyperparameters were found through a small amount of trial and error, i.e., no formal hyperparameter tuning was carried out.

\begin{table}[!htb]
	\centering
	\caption{Toy dataset MIL training hyperparameters.}
	\label{tab:toy_train_hyperparams}
	\centering
	\begin{tabular}{lllll}
		\toprule
		Dataset & LR & WD & Patience & Epochs \\
		\midrule
		Toggle Switch & \num{1e-4} & \num{1e-5} & 20 & 100 \\
		Push Switch & \num{1e-3} & 0 & 30 & 150 \\
		Dial & \num{1e-3} & 0 & 30 & 150 \\
		\bottomrule
	\end{tabular}
\end{table}

In Tables \ref{tab:toy_instance_space_nn_model} to \ref{tab:toy_csc_instance_space_lstm_model} we give the architectures for the MIL models we used in the toy dataset experiments. The models are a combination of fully connected layers (FC) along with different MIL pooling mechanisms. Rectified linear unit (ReLU) activation is applied to the FC hidden layers. We label the layers based on the part of the network they belong to: feature extractor (FE), head network (HN), or pooling (P); see Section \ref{sec:models}. We also indicate the input and output sizes: $b$ x $n$ indicates an input or output where there is a representation of length $n$ for each of the $b$ instances ($b$ is the size of the input bag).

\setlength{\tabcolsep}{3pt}
\begin{table}[!htb]
	\centering
	\begin{minipage}{.49\linewidth}
		\caption{Toy Instance Space NN}
		\vspace{0.7mm}
		\label{tab:toy_instance_space_nn_model}
		\centering
		\begin{tabular}{llll}
			\toprule
			Layer & Type & Input & Output \\
			\midrule
			1 (FE) & FC + ReLU & $b$ x 2 & $b$ x 2 \\
			2 (HN) & FC & $b$ x 2 & $b$ x 1 \\
			3 (P) & mil-sum & $b$ x 1 & 1 \\
			\bottomrule
		\end{tabular}
	\end{minipage}
	\begin{minipage}{.49\linewidth}
		\caption{Toy Embedding Space LSTM}
		\vspace{0.7mm}
		\label{tab:toy_embedding_space_nn_model}
		\centering
		\begin{tabular}{llll}
			\toprule
			Layer & Type & Input & Output \\
			\midrule
			1 (FE) & FC + ReLU & $b$ x 2 & $b$ x 2 \\
			2 (P) & mil-emb-lstm & $b$ x 2 & 2 \\
			3 (HN) & FC & 2 & 1 \\
			\bottomrule
		\end{tabular}
	\end{minipage} 
\end{table}

\begin{table}[!htb]
	\centering
	\begin{minipage}{.49\linewidth}
		\caption{Toy Instance Space LSTM}
		\vspace{0.7mm}
		\label{tab:toy_instance_space_lstm_model}
		\centering
		\begin{tabular}{llll}
			\toprule
			Layer & Type & Input & Output \\
			\midrule
			1 (FE) & FC + ReLU & $b$ x 2 & $b$ x 2 \\
			2 (P-1) & mil-ins-lstm & $b$ x 2 & $b$ x 2 \\
			3 (HN) & FC & $b$ x 2 & $b$ x 1 \\
			4 (P-2) & mil-sum & $b$ x 1 & 1 \\
			\bottomrule
		\end{tabular}
	\end{minipage}
	\begin{minipage}{.49\linewidth}
		\caption{Toy CSC Instance Space LSTM}
		\vspace{0.7mm}
		\label{tab:toy_csc_instance_space_lstm_model}
		\centering
		\begin{tabular}{llll}
			\toprule
			Layer & Type & Input & Output \\
			\midrule
			1 (FE) & FC + ReLU & $b$ x 2 & $b$ x 2 \\
			2 (P-1) & mil-csc-ins-lstm & $b$ x 2 & $b$ x 2 \\
			3 (HN) & FC & $b$ x 2 & $b$ x 1 \\
			4 (P-2) & mil-sum & $b$ x 1 & 1 \\
			\bottomrule
		\end{tabular}
	\end{minipage} 
\end{table}

\subsection{Results}
\label{sec:app_toy_results}

For each of the toy datasets, we generate 5000 random bags with between 10 and 20 instances per bag (uniformly distributed). We use an 80/10/10 dataset split for training, validation, and testing, and repeat our experiments with ten different variations of this split (so in total we have ten repeats of each model type for each dataset). We show results for both return and reward reconstruction for the toy datasets in Table \ref{tab:toy_results}. From these results, we can make several observations. Firstly, as expected, the Instance Space NN architecture only works on the Markovian Toggle Switch dataset, i.e., it fails on the non-Markovian Push Switch and Dial datasets as it is unable to deal with temporal dependencies. We also note that our two proposed architectures (Instance Space LSTM and CSC Instance Space LSTM) outperform the baseline Embedding Space LSTM method on both return and reward, with the CSC Instance Space LSTM model providing the best results overall. Finally, we observe that a better return performance does not always guarantee better reward performance: for the Dial dataset, the Instance Space LSTM makes better return predictions than the CSC Instance Space LSTM model, but worse reward predictions. A similar outcome can be seen for the Embedding Space LSTM and the Instance Space NN on the Push Switch dataset.

\begin{table}[htb!]
	\centering
	\small
	\caption{Toy dataset return (top) and reward (bottom) results. Each measurement is the mean MSE averaged over ten repeats, with the standard errors of the mean also given. Bold entries indicate the best-performing model for each (metric, dataset) pair.}
	\label{tab:toy_results}
	\begin{tabular}{lllll}
		\toprule
		Model & Toggle Switch & Push Switch & Dial & Overall \\
		\midrule
	    Instance Space NN & 0.030 $\pm$ 0.029 & 3.337 $\pm$ 0.054 & 5.489 $\pm$ 0.157 & 2.952 \\
		Embedding Space LSTM & 0.008 $\pm$ 0.002 & 0.663 $\pm$ 0.194 & 0.434 $\pm$ 0.075 & 0.368 \\
		Instance Space LSTM & 0.062 $\pm$ 0.058 & 0.262 $\pm$ 0.154 & \textbf{0.111 $\pm$ 0.014} & 0.145 \\
		CSC Instance Space LSTM & \textbf{0.000 $\pm$ 0.000} & \textbf{0.140 $\pm$ 0.065} & 0.121 $\pm$ 0.043 & \textbf{0.087} \\
		\greyrule
		Instance Space NN & 0.002 $\pm$ 0.002 & 0.086 $\pm$ 0.001 & 0.954 $\pm$ 0.011 & 0.347 \\
		Embedding Space LSTM & 0.003 $\pm$ 0.001 & 0.206 $\pm$ 0.100 & 0.244 $\pm$ 0.077 & 0.151 \\
		Instance Space LSTM & 0.004 $\pm$ 0.004 & 0.021 $\pm$ 0.008 & 0.026 $\pm$ 0.004 & 0.017 \\
		CSC Instance Space LSTM & \textbf{0.000 $\pm$ 0.000} & \textbf{0.012 $\pm$ 0.004} & \textbf{0.022 $\pm$ 0.007} & \textbf{0.011} \\
		\bottomrule
	\end{tabular}
\end{table}

\section{RL Task Details, Data Generation, and Model Hyperparameters}
\label{app:tasks_data_models}

In this section, we give more detail on the reward reconstruction experiments for the RL tasks. First, we give more information about the RL tasks (Section \ref{app:rl_tasks}), then explain how we generated datasets from the tasks (Section \ref{app:dataset_generation}), and give the MIL model architectures and hyperparameters used in the Timer, Moving, Key and Charger tasks (Section \ref{app:mil_rl_models}), and the Lunar Lander task (Section \ref{app:mil_ll}). Finally, we discuss the limitations of our approach to using MIL RM for the Lunar Lander task (Section \ref{app:mil_ll_discussion}).

\subsection{Task Details}
\label{app:rl_tasks}

% All five of our experimental RL tasks are depicted in Figure \ref{fig:env_layouts_two_rows}.

% \begin{figure}[htb!]
%     \centering
%     \includegraphics[width=\textwidth]{figures/env_layouts_two_rows.pdf}
%     \caption{Visualisations of the five non-Markovian RL tasks.}
%     \label{fig:env_layouts_two_rows}
% \end{figure}

The first four tasks are implemented in Python within a common 2D simulator following the OpenAI Gym standard \citep{gym}. The agent's position $x,y$ is moved by one of five discrete actions: up, down, left, right and no-op. In the first four cases, the position is moved by $0.1$ in the specified direction. The motion vector is then corrupted by zero-mean Gaussian noise with a standard deviation of $0.02$ in both $x$ and $y$ and clipped into the bounds $[0,1]^2$. Zones of interest (spawn zones, treasure, key, charger) are specified as rectangles lying within these bounds. At time $t$, the environment state $s_t$ (which is directly observed by the MIL RM models) is the 2D vector of the current position $[x_t,y_t]$; its dynamics are Markovian given the agent's chosen action. The hidden state $h_t$ is the task-specific information that renders the oracle's reward function $R$ Markovian:
\begin{itemize}
    \item \textbf{Timer:} $h_0=0$ and $h_{t+1}=\delta(h_t,s_t,a_t)=h_t+1$; the hidden state simply tracks the current timestep index. Reward is given by\footnote{Note the timestep indices used here, which result from the order in which environment states, hidden states and rewards are computed. At time $t$, the hidden state $h_t$ is first updated to $h_{t+1}$ by $\delta(h_t,s_t,a_t)$, then the reward is computed as $R(s_t,a_t,h_{t+1})$, and finally the environment state is updated to $s_{t+1}$ by $D(s_t,a_t)$.}
    $$
    R(s_t,a_t,h_{t+1})=\text{in\_treasure}(s_t)\cdot
    \left\{\begin{array}{ll} 
    -1 & \text{if }h_{t+1}\leq 50,\\
    +1 & \text{otherwise,}
    \end{array}\right.
    $$
    where $\text{in\_treasure}([x_t,y_t])=1$ if $0.4\leq x_t\leq 0.6$ and $0.7\leq y_t\leq 0.9$, and $0$ otherwise.
    \item \textbf{Moving:} $h_0=[0.4,-0.02]$, the initial horizontal position (left edge) and velocity of the moving treasure rectangle. Hidden state dynamics encode the left-right oscillation:
    $$
    h_{t+1}=\delta(h_t,s_t,a_t)=\left[\ h_t^0+h_t^1,\ 
    \left\{\begin{array}{ll}
    h_t^1&\text{if }0<(h_t^0+h_t^1)<0.8,\\
    -h_t^1&\text{otherwise}
    \end{array}\right.\right].
    $$
    Reward is given by $R([x_t,y_t],a_t,h_{t+1})=1$ if $h_{t+1}\leq x_t\leq (h_{t+1}+0.2)$ and $0.7\leq y_t\leq 0.9$, and $0$ otherwise.
    
    \item \textbf{Key:} $h_0=0$, indicating that the agent initialises without the key. The key collection dynamics are encoded by
    $$h_{t+1}=\delta(h_t,[x_t,y_t],a_t)=\left\{\begin{array}{ll}
    1&\text{if }0.4\leq x_t\leq 0.6\text{ and }0.1\leq y_t\leq 0.3,\\
    h_t&\text{otherwise.}
    \end{array}\right.
    $$
    Reward is given by $R(s_t,a_t,h_{t+1})=\text{in\_treasure}(s_t)\cdot h_{t+1}$, where the $\text{in\_treasure}$ function is the same as in the Timer task.
    
    \item \textbf{Charger:} $h_0=0$, indicating an initial charge level of zero. The charging dynamics are encoded by
    $$h_{t+1}=\delta(h_t,[x_t,y_t],a_t)=\left\{\begin{array}{ll}
    \min(h_t+0.02,1)&\text{if }y_t\leq 0.3,\\
    h_t&\text{otherwise.}
    \end{array}\right.
    $$
    Reward is given identically to the Key task, $R(s_t,a_t,h_{t+1})=\text{in\_treasure}(s_t)\cdot h_{t+1}$.
\end{itemize}

The \textbf{Lunar Lander} task is a modified version of the \textsc{LunarLanderContinuous-v2} baseline included as standard in the OpenAI Gym library \cite{gym}. We leave the state and action spaces unmodified. The 8D state vector is $[x,y,v^x,v^y,\theta,\dot{\theta},c^l,c^r]$, where $x,y$ and $v^x,v^y$ are the landing craft's horizontal and vertical positions and velocities, $\theta$ and $\dot{\theta}$ are its angle from vertical and angular velocity, and $c^l,c^r$ are two binary contact detectors indicating whether the left and right landing legs are in contact with the ground. The 2D continuous action $[u^m,u^s]$ is a pair of throttle values for two engines: main $u^m$ and side $u^s$. We also retain the default initialisation conditions (the lander spawns in a narrow zone above the landing pad, with slightly-randomised orientation and velocities), the automatic termination of episodes when $|x|$ exceeds $1$ (i.e., when the lander leaves the rendered screen area), and the physics that determine how the lander responds to engine activations. However, we replace the standard reward function with an oracle that rewards the agent for landing on the pad for up to $50$ timesteps, and then taking off again to hover within a target zone until an episode time limit ($T=500$) is reached. Rendering this two-stage objective Markovian requires a hidden state $h_t$ that tracks the number of timesteps spent on the pad so far. Formally, reward is given by
$$
R(s_t,a_t,h_{t+1})=
\left\{\begin{array}{ll} 
R_\text{pad}(s_t)+R_\text{shaping}(s_t,0) & \text{if }h_{t+1}< 50,\\
R_\text{no\_contact}(s_t)+R_\text{hover}(s_t)+R_\text{shaping}(s_t,1) & \text{otherwise,}
\end{array}\right.
$$
where $R_\text{pad}$ rewards the agent for being central with both legs on the ground (i.e., on the pad),
$$
R_\text{pad}([x_t,y_t,v^x_t,v^y_t,\theta_t,\dot{\theta}_t,c^l_t,c^r_t])=\left\{\begin{array}{ll} 
1 & \text{if }-0.2\leq x_t\leq 0.2\text{ and }c^l_t=1\text{ and }c^r_t=1,\\
0 & \text{otherwise,}
\end{array}\right.
$$
$R_\text{no\_contact}$ rewards breaking leg-ground contact,
$$
R_\text{no\_contact}([x_t,y_t,v^x_t,v^y_t,\theta_t,\dot{\theta}_t,c^l_t,c^r_t])=\left\{\begin{array}{ll} 
1 & \text{if }c^l_t=0\text{ and }c^r_t=0,\\
0 & \text{otherwise,}
\end{array}\right.
$$
$R_\text{hover}$ rewards aerial positions in a target zone above the pad,
$$
R_\text{hover}([x_t,y_t,v^x_t,v^y_t,\theta_t,\dot{\theta}_t,c^l_t,c^r_t])=\left\{\begin{array}{ll} 
1 & \text{if }-0.5\leq x_t\leq 0.5\text{ and }0.75\leq y_t\leq 1.25,\\
0 & \text{otherwise,}
\end{array}\right.
$$
and $R_\text{shaping}$ promotes slow, stable, central flight towards a target vertical position $y_\text{target}$,
\begin{multline*}
R_\text{shaping}([x_t,y_t,v^x_t,v^y_t,\theta_t,\dot{\theta}_t,c^l_t,c^r_t], y_\text{target})=\\
0.1\times\max\left(2 - \left(\sqrt{(x_t)^2+(y_t-y_\text{target})^2}+\sqrt{(v^x_t)^2+(v^y_t)^2}+|\theta_t|+|\dot{\theta}_t|\right),\ 0\right).
\end{multline*}
The hidden state dynamics are
$$h_{t+1}=\left\{\begin{array}{ll}
\min(h_t+1,50)&\text{if }R_\text{pad}(s_t)=1,\\
h_t&\text{otherwise,}
\end{array}\right.
$$
with $h_0=0$.

% To convert Lunar Lander into a fixed-length episodic task, we disabled a default condition that terminates the episode immediately after a landing, crash or out-of-bounds event.

\subsection{MIL Dataset Generation}
\label{app:dataset_generation}

We obtain datasets of several thousand trajectories per task, containing a wide distribution of outcomes and return values, as follows. For each task $k$, we define a discrete \textit{trajectory classifier} function $C_k:\Xi\rightarrow\mathcal{C}_k$ and a limit $p_k$ on the proportion of trajectories in the dataset that are allowed to map to each class in $\mathcal{C}_k$. These are given as follows:
\begin{itemize}
    \item \textbf{Timer:} $\mathcal{C}_\text{timer}=\text{num\_neg}\times\text{num\_pos}$, where $\text{num\_neg}=\{0..50\}$ counts the number of timesteps the agent spends in the treasure while its reward is negative ($t\leq 50$), and $\text{num\_pos}=\{0..50\}$ counts the number while the reward is positive ($t>50$). The number of classes is $|\mathcal{C}_\text{timer}|=51^2=2601$ and the per-class limit is $p_\text{timer}=0.002$.
    \item \textbf{Moving:} $\mathcal{C}_\text{moving}=\text{num\_treasure}$, where $\text{num\_treasure}=\{0..100\}$ counts the timesteps spent in the treasure. $|\mathcal{C}_\text{moving}|=101$ and $p_\text{moving}=0.05$.
    \item \textbf{Key:} $\mathcal{C}_\text{key}=\{\text{no\_key},\text{key\_no\_treasure},\text{treasure}\}$, where the class is $\text{no\_key}$ if the key is not collected, $\text{key\_no\_treasure}$ if the key is collected but the treasure is not reached, and $\text{treasure}$ if the treasure is reached after collecting the key. $|\mathcal{C}_\text{key}|=3$ and $p_\text{key}$ is defined on a per-class basis: $0.25$ for $\text{no\_key}$ and $\text{key}$, and $0.5$ for $\text{treasure}$.
    \item \textbf{Charger:} $\mathcal{C}_\text{charger}=\text{num\_treasure}\times \text{charge\_bin}$, where $\text{num\_treasure}=\{0..100\}$ counts the timesteps spent in the treasure and $\text{charge\_bin}=\{1..20\}$ is a binned representation of the \textit{mean} charge level when in the treasure (e.g., $0.0$ maps to bin $1$, $0.48$ to bin $10$, $0.96$ to bin $20$). $|\mathcal{C}_\text{charger}|=2020$ and $p_\text{charger}=0.002$.
    \item \textbf{Lunar Lander:} $\mathcal{C}_\text{lunar}=\text{pad\_bin}\times\text{take\_off}\times\text{hover\_bin}$, where $\text{pad\_bin}=\{0,\{1..49\},50+\}$ is a binned representation of the number of timesteps spent on the landing pad (i.e., zero, fewer than $50$ or at least $50$), $\text{take\_off}=\{0,1\}$ is a binary indicator of whether the lander takes off again after being on the pad, and $\text{hover\_bin}=\{0,\{1..19\},20+\}$ is a binned representation of the number of timesteps spent in the hover zone after being on the pad.\footnote{If the lander reaches the target of $50$ timesteps on the pad, the time in the hover zone is measured from this point onwards. Otherwise, it is measured from the first timestep that the lander leaves the pad.} $|\mathcal{C}_\text{lunar}|=18$, of which $9$ are actually realisable (e.g., the lander cannot take off from the pad if it never reached it in the first place) and $p_\text{lunar}=0.2$.
\end{itemize}
For $k\in\{\text{timer},\text{moving},\text{key},\text{charger}\}$, a dataset $\textbf{X}_k$, is assembled iteratively. On each iteration, we generate a length-100 trajectory, $\xi$, by sampling agent actions uniform-randomly from the action space (up, down, left, right, no-op) and running them through the simulator. Once the trajectory is complete, we evaluate its class $C_k(\xi)$. If there are already at least $p_k\times 5000$ trajectories in $\textbf{X}_k$ with this class, $\xi$ is discarded. Otherwise, it is added to $\textbf{X}_k$. This process repeats until $|\textbf{X}_k|=5000$.

The state-action space for Lunar Lander is too large for a random generate-and-select algorithm to terminate in any reasonable time. Instead, we recycle the length-$500$ trajectories generated as a by-product of training the oracle-based RL baselines (black curves in Figure \ref{fig:rl_training_results}, plus six more runs not included in the figure). Starting from a bank of $12000$ trajectories, filtering based on the threshold $p_\text{lunar}=0.2$ yields a final dataset $\textbf{X}_\text{lunar}$ with $9762$ trajectories. Although this approach of relying on oracle-trained agents to generate data may initially appear to ``put the cart before the horse'', we suggest that it provides a valuable test of the ability of our MIL models to learn from goal-directed (c.f. random) trajectories, and thus is a step closer to the online bootstrapping approach of simultaneous RM and RL, which we aim to tackle in future work (see Section \ref{sec:future_work}).

\subsection{MIL Models and Hyperparameters}
\label{app:mil_rl_models}

The MIL model training on the Timer, Moving, Key, and Charger tasks used the same process as for the toy model training (see Section \ref{sec:app_toy_models}). However, we also applied dropout (DO) in these models. We give the MIL training hyperparameters for each of these tasks in Table \ref{tab:mil_rl_train_hyperparams}. Again, the hyperparameters were found through a small amount of trial and error, i.e., no formal hyperparameter tuning was carried out.

\begin{table}[!htb]
	\centering
	\caption{Timer, Moving, Key, and Charger task training hyperparameters.}
	\label{tab:mil_rl_train_hyperparams}
	\centering
	\begin{tabular}{llllll}
		\toprule
		Dataset & LR & WD & DO & Patience & Epochs \\
		\midrule
		Timer & \num{5e-4} & 0 & 0.1 & 50 & 250 \\
		Moving & \num{5e-4} & 0 & 0.1 & 50 & 250 \\
		Key & \num{5e-4} & 0 & 0.1 & 30 & 150 \\
		Charger & \num{5e-4} & 0 & 0.1 & 50 & 250 \\
		\bottomrule
	\end{tabular}
\end{table}

In Tables \ref{tab:mil_rl_instance_space_nn_model} to \ref{tab:mil_rl_csc_instance_space_lstm_model} we give the architectures for the MIL models we used in the Timer, Moving, Key, and Charger tasks. As in the toy dataset experiments, the models are a combination of fully connected layers (FC) along with different MIL pooling mechanisms. Rectified linear unit (ReLU) activation is applied to the FC hidden layers. Again, we label the layers based on the part of the network they belong to: feature extractor (FE), head network (HN), or pooling (P); see Section \ref{sec:models}. We also indicate the input and output sizes: $b$ x $n$ indicates an input or output where there is a representation of length $n$ for each of the $b$ instances ($b$ is the size of the input bag). Input features were normalised using mean/standard deviation scaling.

\setlength{\tabcolsep}{3pt}
\begin{table}[!htb]
	\centering
	\begin{minipage}{.49\linewidth}
		\caption{RL Instance Space NN}
		\vspace{0.7mm}
		\label{tab:mil_rl_instance_space_nn_model}
		\centering
		\begin{tabular}{llll}
			\toprule
			Layer & Type & Input & Output \\
			\midrule
			1 (FE-1) & FC + ReLU + DO & $b$ x 2 & $b$ x 64 \\
			2 (FE-2) & FC + ReLU + DO & $b$ x 64 & $b$ x 32 \\
			3 (FE-3) & FC + ReLU + DO & $b$ x 32 & $b$ x 32 \\
			4 (HN-1) & FC + ReLU + DO & $b$ x 32 & $b$ x 32 \\
			5 (HN-2) & FC + ReLU + DO & $b$ x 32 & $b$ x 16 \\
			6 (HN-3) & FC & $b$ x 16 & $b$ x 1 \\
			7 (P) & mil-sum & $b$ x 1 & 1 \\
			\bottomrule
		\end{tabular}
	\end{minipage}
	\begin{minipage}{.49\linewidth}
		\caption{RL Embedding Space LSTM}
		\vspace{0.7mm}
		\label{tab:mil_rl_embedding_space_nn_model}
		\centering
		\begin{tabular}{llll}
			\toprule
			Layer & Type & Input & Output \\
			\midrule
			1 (FE-1) & FC + ReLU + DO & $b$ x 2 & $b$ x 64 \\
			2 (FE-2) & FC + ReLU + DO & $b$ x 64 & $b$ x 32 \\
			3 (FE-3) & FC + ReLU + DO & $b$ x 32 & $b$ x 32 \\
			4 (P) & mil-emb-lstm & $b$ x 32 & 2 \\
			5 (HN-1) & FC + ReLU + DO & 2 & 32 \\
			6 (HN-2) & FC + ReLU + DO & 32 & 16 \\
			7 (HN-3) & FC & 16 & 1 \\
			\bottomrule
		\end{tabular}
	\end{minipage} 
\end{table}

\vspace{-0.2cm}
\begin{table}[!htb]
	\centering
	\begin{minipage}{.49\linewidth}
		\caption{RL Instance Space LSTM}
		\vspace{0.7mm}
		\label{tab:mil_rl_instance_space_lstm_model}
		\centering
		\begin{tabular}{llll}
			\toprule
			Layer & Type & Input & Output \\
			\midrule
			1 (FE-1) & FC + ReLU + DO & $b$ x 2 & $b$ x 64 \\
			2 (FE-2) & FC + ReLU + DO & $b$ x 64 & $b$ x 32 \\
			3 (FE-3) & FC + ReLU + DO & $b$ x 32 & $b$ x 32 \\
			4 (P-1) & mil-ins-lstm & $b$ x 32 & $b$ x 2 \\
		    5 (HN-1) & FC + ReLU + DO & $b$ x 2 & $b$ x 32 \\
			6 (HN-2) & FC + ReLU + DO & $b$ x 32 & $b$ x 16 \\
			7 (HN-3) & FC & $b$ x 16 & $b$ x 1 \\
			8 (P-2) & mil-sum & $b$ x 1 & 1 \\
			\bottomrule
		\end{tabular}
	\end{minipage}
	\begin{minipage}{.49\linewidth}
		\caption{RL CSC Instance Space LSTM}
		\vspace{0.7mm}
		\label{tab:mil_rl_csc_instance_space_lstm_model}
		\centering
		\begin{tabular}{llll}
			\toprule
			Layer & Type & Input & Output \\
			\midrule
			1 (FE-1) & FC + ReLU + DO & $b$ x 2 & $b$ x 64 \\
			2 (FE-2) & FC + ReLU + DO & $b$ x 64 & $b$ x 32 \\
			3 (FE-3) & FC + ReLU + DO & $b$ x 32 & $b$ x 32 \\
			4 (P-1) & mil-csc-ins-lstm & $b$ x 32 & $b$ x 2 \\
		    5 (HN-1) & FC + ReLU + DO & $b$ x 34 & $b$ x 32 \\
			6 (HN-2) & FC + ReLU + DO & $b$ x 32 & $b$ x 16 \\
			7 (HN-3) & FC & $b$ x 16 & $b$ x 1 \\
			8 (P-2) & mil-sum & $b$ x 1 & 1 \\
			\bottomrule
		\end{tabular}
	\end{minipage} 
\end{table}

\subsection{Lunar Lander MIL Models and Hyperparameters}
\label{app:mil_ll}

There are several differences between the MIL training process for the Lunar Lander task and the other four RL tasks (see Appendix \ref{app:mil_rl_models}). Firstly, the reward targets (and as such, return targets) were scaled down by a factor of 100 in order to avoid extremely large gradients from high prediction targets. For example, a trajectory with an original return of 700 would have a scaled return of 7. Secondly, the input data was scaled linearly between -0.5 and 0.5 (using the minimum and maximum range of each feature). This was found to give more consistent feature ranges than mean/standard deviation scaling as was used in the other tasks (this was due to large outliers in certain features, e.g., the rotational features were largely clustered around 0, but had extreme values up to $\pm$ 90). We give the training hyperparameters for the lunar lander environment in Table \ref{tab:mil_rl_train_hyperparams}. Again, no formal hyperparameter tuning was carried out, so better performance of these models could potentially be achieved with better parameters (including shorter training times with a higher learning rate).

\begin{table}[!htb]
	\centering
	\caption{Lunar Lander MIL training hyperparameters.}
	\label{tab:mil_rl_train_hyperparams}
	\centering
	\begin{tabular}{llllll}
		\toprule
		Dataset & LR & WD & DO & Patience & Epochs \\
		\midrule
		Lunar Lander & \num{1e-4} & 0 & 0 & 30 & 200 \\
		\bottomrule
	\end{tabular}
\end{table}

We used similar architectures to the other four RL tasks (see Appendix \ref{app:mil_rl_models}), but with larger layers; see Tables \ref{tab:mil_ll_instance_space_nn_model} through \ref{tab:mil_ll_csc_instance_space_lstm_model}. However, the depth of the models remained the same, as did the size of the hidden state embedding. Also note the addition of the Leaky ReLU activation function in the head networks, which we discuss further in Appendix \ref{app:mil_ll_discussion}.

\setlength{\tabcolsep}{3pt}
\begin{table}[!htb]
	\centering
	\begin{minipage}{.49\linewidth}
		\caption{Lunar Lander Instance Space NN}
		\vspace{0.7mm}
		\label{tab:mil_ll_instance_space_nn_model}
		\centering
		\begin{tabular}{llll}
			\toprule
			Layer & Type & Input & Output \\
			\midrule
			1 (FE-1) & FC + ReLU + DO & $b$ x 2 & $b$ x 128 \\
			2 (FE-2) & FC + ReLU + DO & $b$ x 128 & $b$ x 64 \\
			3 (FE-3) & FC + ReLU + DO & $b$ x 64 & $b$ x 64 \\
			4 (HN-1) & FC + ReLU + DO & $b$ x 64 & $b$ x 64 \\
			5 (HN-2) & FC + ReLU + DO & $b$ x 64 & $b$ x 32 \\
			6 (HN-3) & FC + Leaky ReLU & $b$ x 32 & $b$ x 1 \\
			7 (P) & mil-sum & $b$ x 1 & 1 \\
			\bottomrule
		\end{tabular}
	\end{minipage}
	\begin{minipage}{.49\linewidth}
		\caption{Lunar Lander Emb. Space LSTM}
		\vspace{0.7mm}
		\label{tab:mil_ll_embedding_space_nn_model}
		\centering
		\begin{tabular}{llll}
			\toprule
			Layer & Type & Input & Output \\
			\midrule
			1 (FE-1) & FC + ReLU + DO & $b$ x 2 & $b$ x 128 \\
			2 (FE-2) & FC + ReLU + DO & $b$ x 128 & $b$ x 64 \\
			3 (FE-3) & FC + ReLU + DO & $b$ x 64 & $b$ x 64 \\
			4 (P) & mil-emb-lstm & $b$ x 64 & 2 \\
			5 (HN-1) & FC + ReLU + DO & 2 & 64 \\
			6 (HN-2) & FC + ReLU + DO & 64 & 32 \\
			7 (HN-3) & FC + Leaky ReLU & 32 & 1 \\
			\bottomrule
		\end{tabular}
	\end{minipage} 
\end{table}

\vspace{-0.2cm}
\begin{table}[!htb]
	\centering
	\begin{minipage}{.49\linewidth}
		\caption{Lunar Lander Ins. Space LSTM}
		\vspace{0.7mm}
		\label{tab:mil_ll_instance_space_lstm_model}
		\centering
		\begin{tabular}{llll}
			\toprule
			Layer & Type & Input & Output \\
			\midrule
			1 (FE-1) & FC + ReLU + DO & $b$ x 2 & $b$ x 128 \\
			2 (FE-2) & FC + ReLU + DO & $b$ x 128 & $b$ x 64 \\
			3 (FE-3) & FC + ReLU + DO & $b$ x 64 & $b$ x 64 \\
			4 (P-1) & mil-ins-lstm & $b$ x 64 & $b$ x 2 \\
		    5 (HN-1) & FC + ReLU + DO & $b$ x 2 & $b$ x 64 \\
			6 (HN-2) & FC + ReLU + DO & $b$ x 64 & $b$ x 32 \\
			7 (HN-3) & FC + Leaky ReLU & $b$ x 32 & $b$ x 1 \\
			8 (P-2) & mil-sum & $b$ x 1 & 1 \\
			\bottomrule
		\end{tabular}
	\end{minipage}
	\begin{minipage}{.49\linewidth}
		\caption{Lunar Lander CSC Ins. Space LSTM}
		\vspace{0.7mm}
		\label{tab:mil_ll_csc_instance_space_lstm_model}
		\centering
		\begin{tabular}{llll}
			\toprule
			Layer & Type & Input & Output \\
			\midrule
			1 (FE-1) & FC + ReLU + DO & $b$ x 2 & $b$ x 128 \\
			2 (FE-2) & FC + ReLU + DO & $b$ x 128 & $b$ x 64 \\
			3 (FE-3) & FC + ReLU + DO & $b$ x 64 & $b$ x 64 \\
			4 (P-1) & mil-csc-ins-lstm & $b$ x 64 & $b$ x 2 \\
		    5 (HN-1) & FC + ReLU + DO & $b$ x 66 & $b$ x 64 \\
			6 (HN-2) & FC + ReLU + DO & $b$ x 64 & $b$ x 32 \\
			7 (HN-3) & FC + Leaky ReLU & $b$ x 32 & $b$ x 1 \\
			8 (P-2) & mil-sum & $b$ x 1 & 1 \\
			\bottomrule
		\end{tabular}
	\end{minipage} 
\end{table}

\subsection{Lunar Lander MIL Discussion}
\label{app:mil_ll_discussion}

In Table \ref{tab:rl_results}, we present results for Lunar Lander using only the top five performing models by reward MSE (50\% of all models). In this section, we discuss why this is the case. 

When training the Lunar Lander architectures with linear activation functions in the head networks, we found that the models were struggling to learn the correct return predictions as they were making negative reward predictions. We know \textit{a priori} that the Lunar Lander task only has positive rewards, therefore we added a Leaky ReLU activation (with a negative slope of \num{1e-6}) to each architecture's head network to encourage positive predictions. This led to an immediate improvement in performance for all model architecture (excluding the Instance Space NN baseline, but this was expected to fail on this task as it cannot model temporal dependencies).

However, we found that a proportion of model initialisations remained unable to overcome a certain local minimum during training (corresponding to a return prediction MSE of around $2.1$). In other cases, after this threshold was passed, the model performance would rapidly improve. Note this problem occurred in each of the LSTM-based models (the Instance Space NN architecture was never observed to pass this threshold). Therefore, we focus our evaluation on the top 50\% of models, which is equivalent to discarding the worse-performing models, the majority of which had not passed the problematic threshold during training.

\begin{figure}[htb!]
    \centering
    \includegraphics[width=\textwidth]{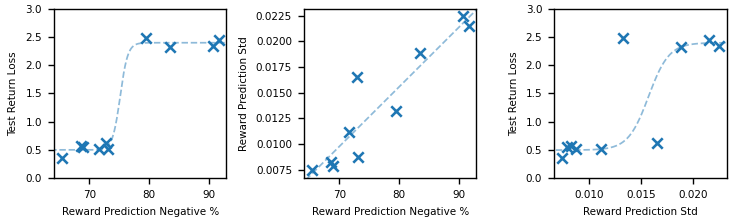}
    \caption{An analysis of the negative reward prediction of the ten Lunar Lander CSC Instance Space LSTM models trained in this work. Note we observed similar trends for the Embedding Space LSTM and Instance Space LSTM models. \textbf{Left:} Models that are able to cross the return loss threshold of 2.1 make fewer negative reward predictions (given here as a percentage of all reward predictions) than those that cannot. \textbf{Middle:} As the number of negative reward predictions increases, so too does the standard deviation of the reward predictions. In order to sum to the correct return predictions, the positive reward predictions must increase to compensate for the negative reward predictions, leading to a larger variance, and ultimately worse reward prediction. \textbf{Right:} The standard deviation of the reward predictions also highlights the loss threshold of 2.1, although not as clearly as the percentage of negative reward predictions.}
    \label{fig:csc_negative_reward_pred_analysis}
\end{figure}

We investigate this training issue further in Figure \ref{fig:csc_negative_reward_pred_analysis}, focusing specifically on the CSC Instance Space LSTM models. Although the Leaky ReLU activation encourages models to make positive predictions, we can see that it does not prevent them entirely. Furthermore, the models that are not able to cross the return threshold of $2.1$ tend to output a greater proportion of negative reward predictions. We thus hypothesise that a better mechanism for preventing negative reward prediction would increase the chance that a given model training run achieves a return loss of less than $2.1$. Below we list several such mechanisms as alternatives to our current Leaky ReLU approach.

\paragraph{Replace Leaky ReLU with ReLU} \hspace{-0.8em} One option to remove negative reward predictions entirely is to use ReLU rather than Leaky ReLU in the final activation function of the head networks. However, in the case that all the reward predictions are negative, the gradient of the network will be zero, so no learning can take place. The ability of the network to learn is entirely dependent on achieving at least one positive prediction in order to generate non-zero gradients, which is determined by the initial network weights. This could potentially be overcome with different network initialisation approaches, or by simply discarding training runs that fail to begin to learn.

\paragraph{Replace Leaky ReLU with Sigmoid} \hspace{-0.8em} Instead of using a Leaky ReLU or ReLU activation in the head network, the Sigmoid activation function could be used. This would overcome the gradient issue presented with the use of ReLU, and would ensure that no negative rewards are predicted. However, this would require \textit{a priori} knowledge of the maximum reward target in order to scale the network outputs correctly, and the non-linear activation could make accurate prediction of rewards more difficult in some cases.

\paragraph{Remove target normalisation} \hspace{-0.8em} A further approach to reduce the number of negative reward predictions would be to remove, or at least reduce, the reward target scaling. As discussed above, this was initially included to avoid very large gradients. However, reducing this scaling would move the average reward prediction away from zero, potentially reducing the number of negative predictions. Care would have to be taken to not reintroduce the large gradient problem.

\paragraph{Regularise reward prediction variance} \hspace{-0.8em} A final alternative approach could be to introduce an additional loss term that encourages the reward predictions to have low variance. This would deter a large range of reward predictions, leading indirectly to fewer negative reward predictions (see Figure \ref{fig:csc_negative_reward_pred_analysis}). However, this approach is only applicable to the LSTM models that produce instance predictions, i.e., it cannot be used with the Embedding Space LSTM network as that architecture does not produce reward predictions during training. Furthermore, this would result in an additional hyperparameter that requires tuning (a coefficient for the new loss term).

\section{Further Details on RL Agent Training}
\label{app:rl_training_appendix}

Adopting OpenAI Gym terminology \citep{gym}, non-Markovian reward functions (both ground truth oracles and learnt LSTM-based models) are implemented as \textit{wrappers} on rewardless base environments. The role of a wrapper is to track the hidden state of either the oracle or the LSTM throughout an episode and use this to compute rewards to return to the agent. In the ``Oracle (without hidden state)'' baseline, we return the raw environment state (e.g., the 2D position $[x_t,y_t]$) to the agent unmodified. Otherwise, we concatenate the post-update hidden state $h_{t+1}$ onto the end of the environment state, thereby expanding the state space from the agent's perspective and making rewards Markovian. 

This wrapper-based approach allows us to use a completely vanilla RL algorithm. For the 2D navigation tasks, we use a Deep Q-Network (DQN) agent \citep{mnih2015human} with the double Q-learning trick \citep{van2016deep} enabled. For Lunar Lander, which has a continuous action space, we use Soft Actor-Critic (SAC) \citep{haarnoja2018soft}. In both cases we use a value network with ReLU activation functions, which is updated on every timestep by sampling batches of size $128$ from a replay buffer. Bellman updates use a discount factor of $\gamma=0.99$ and are implemented by the Adam optimiser with a learning rate of $1\text{e}^{-3}$. A target network tracks the primary one by Polyak averaging of parameters with a coefficient of $5\text{e}^{-3}$ per timestep. Additional hyperparameters are given below:
\begin{itemize}
    \item \textbf{2D tasks (DQN):} number of training episodes $=400$; replay buffer capacity $=5\text{e}^4$; value network hidden layer sizes $=[256,128,64]$; policy greediness $\epsilon$ linearly decayed from $1$ to $0.05$ over the first $200$ episodes and held constant thereafter.
    \item \textbf{Lunar Lander (SAC):} number of training episodes $=800$; replay buffer capacity $=1\text{e}^5$; value/policy network hidden layer sizes $=[256,256]$; policy entropy regularisation coefficient $\alpha=0.2$; policy updates with Adam optimiser (learning rate $1\text{e}^{-4}$).
\end{itemize}

\section{Trajectory Probing for Other RL Tasks}
\label{app:other_task_interpretability}

\begin{figure}[htb!]
    \centering
    \includegraphics[width=\textwidth]{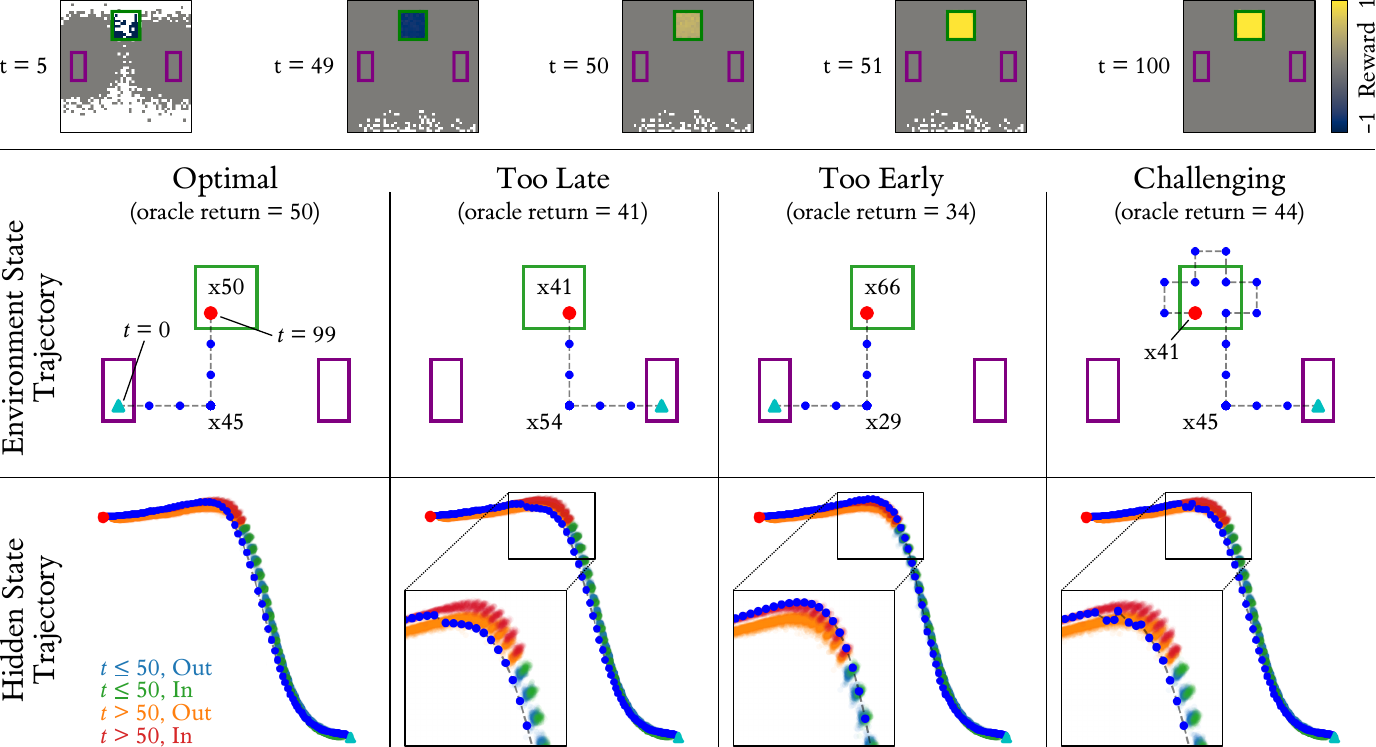}
    \caption{
        Timer task trajectory probing. \\
        \textbf{Top}: Predicted reward with respect to time and position. We observe that the model has correctly captured the transition from negative to positive reward in the treasure region at $t=50$, with no reward outside of this region. Although the reward is positive at $t=50$, the model is uncertain at this point, i.e., the transition from negative to positive reward happens over two timesteps rather than one. \\
        \textbf{Middle/bottom}: Four trajectory probes demonstrating the model's hidden state transitions. ``\texttt{x}$n$'' labels indicate the agent remaining in a position for $n$ timesteps. \\
        \textit{Optimal}: The agent moves into the treasure region at $t=50$ and remains there, receiving the maximum possible reward.\\
        \textit{Too Late}: The agent moves into the treasure region a while after the treasure has already become positive, i.e., it is missing out on reward by not being in the region for as long as possible. This is reflected in the hidden state plot, where the state transitions from the orange to the red region after the $t=50$ boundary. \\
        \textit{Too Early}: The agent moves into the treasure region before the treasure becomes positive, therefore, while it earns the maximum amount of positive reward, it also earns negative reward, leading to a sub-optimal result. \\
        \textit{Challenging}: The agent moves into the treasure region at the correct time, but proceeds to jump in and out of the treasure region before settling, leading to lost reward. The hidden state trajectories somewhat mimic this movement by transitioning between the orange and red regions, although the jumps are less clear near to the $t=50$ transition point, suggesting the model is uncertain at this point.
    }
    \label{fig:timer_interpretability}
\end{figure}

\begin{figure}[htb!]
    \centering
    \includegraphics[width=\textwidth]{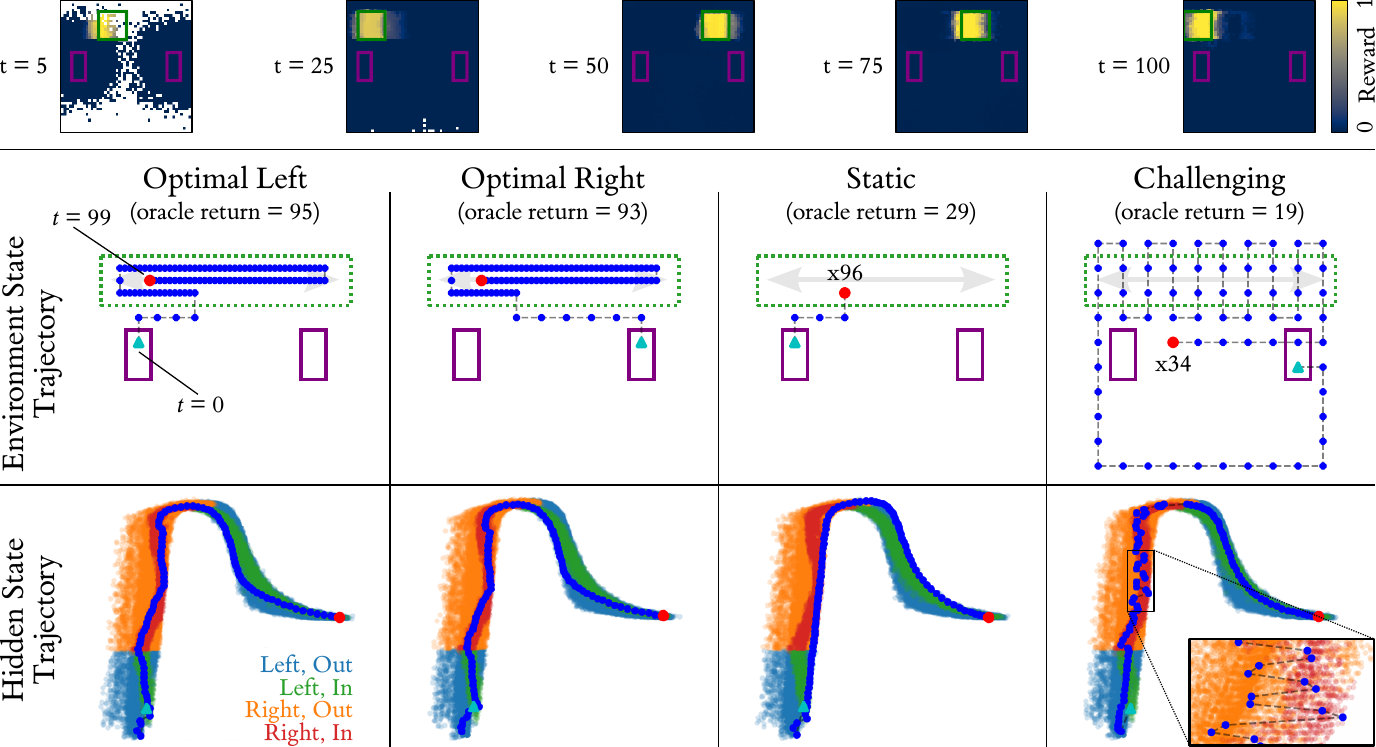}
    \caption{
        Moving task trajectory probing. \\
        \textbf{Top}: The predicted reward with respect to time (and thus implicitly, the position of the treasure region). The model has learnt to track the treasure region as it moves, although there is noise around the left and right edges of the region, highlighting the difficulty of recovering the treasure region's horizontal position exactly. \\
        \textbf{Middle/bottom}: Four trajectory probes showing the model's hidden state transitions. Note the green dotted region indicates the overall boundary of the treasure region, i.e., the treasure lies somewhere within that boundary, with its true horizontal position dependent on time. ``\texttt{x}$n$'' labels indicate the agent remaining in a position for $n$ timesteps. \\
        \textit{Optimal Left}: The agent moves within the treasure region as quickly as possible and then moves with the treasure (keeping within it) for the remainder of the episode. The hidden state trajectories follow the ``In'' regions (green and red). \\
        \textit{Optimal Right}: A similar optimal probe where the agent starts from the right-hand spawn zone rather than the left. In this case, the agent has further to move before moving into the treasure region, leading to slightly less overall reward than when it starts from the left. \\
        \textit{Static}: Rather than moving with the treasure region, the agent stays still and allows the treasure to pass over it, gaining reward for some timesteps but not others. We observe that the hidden state trajectory also reflects this \textemdash{} the state transitions between the ``In'' and ``Out'' regions.\\
        \textit{Challenging}: The agent takes a while to move towards the treasure region, and then passes in and out of the boundary in which the treasure resides, picking up some reward. We can see from the hidden state trajectory that this motion is captured in the hidden states \textemdash{} the trajectory transitions between the orange and red regions as the agent passes back and forth through the treasure region.
    }
    \label{fig:moving_interpretability}
\end{figure}

\begin{figure}[htb!]
    \centering
    \includegraphics[width=\textwidth]{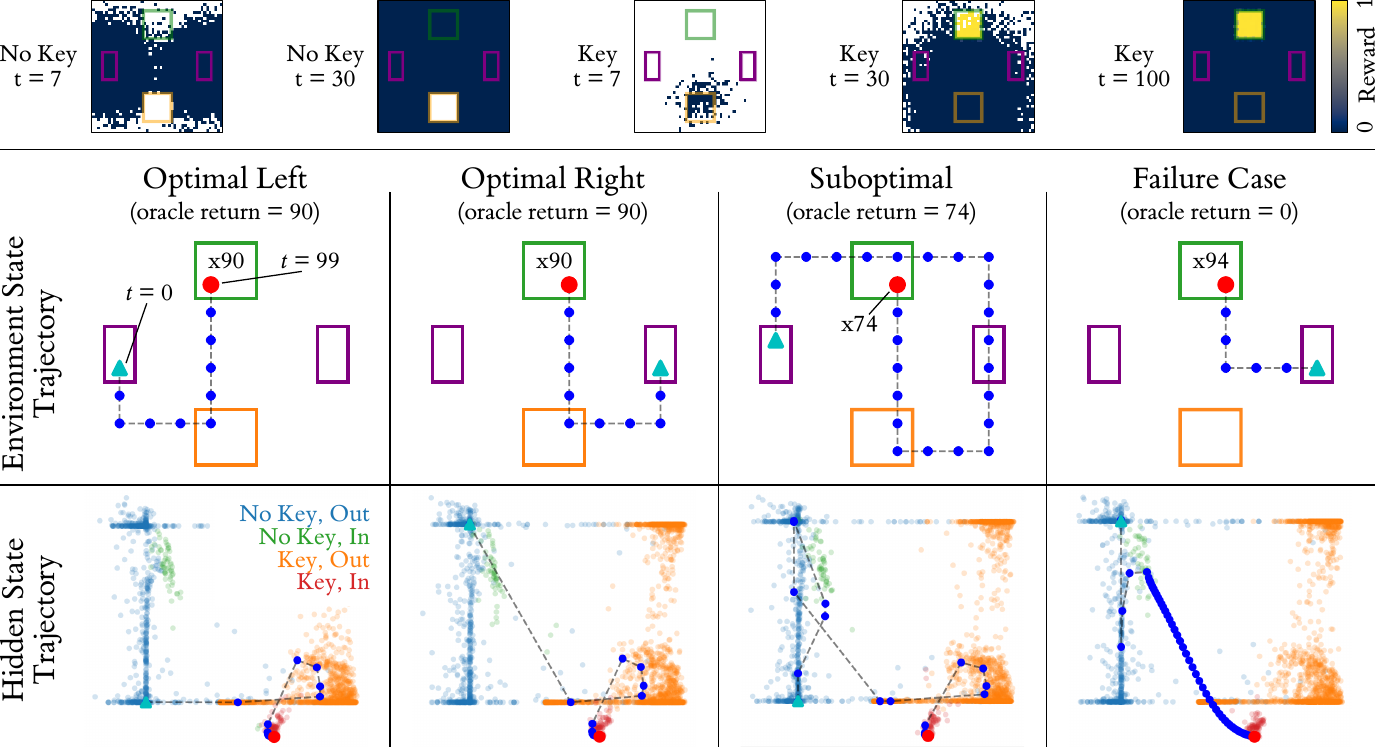}
    \caption{
        Key task trajectory probing. \\
        \textbf{Top}: The predicted reward with respect to time and collection of the key. The model has learnt to only give reward when the agent is in the treasure region after collecting the key. \\
        \textbf{Middle/bottom}: Four trajectory probes showing the model's hidden state transitions. ``\texttt{x}$n$'' labels indicate the agent remaining in a position for $n$ timesteps. \\
        \textit{Optimal Left}: The agent collects the key as quickly as possible, and then proceeds to move into the treasure region and wait there until the end of the episode. The hidden state trajectory shows two notable transitions, first when the key is collected (blue to orange) and then when the agent enters the treasure region (orange to red). \\
        \textit{Optimal Right}: A similar optimal policy, but from the right-hand spawn zone rather than the left. Note the hidden state trajectory is very similar, apart from the initial hidden state, which is at the opposite side of the blue region, representing the different spawn position. \\
        \textit{Suboptimal}: The agent passes through the treasure region before collecting the key, and then follows an optimal policy. We observe a transition in the hidden state trajectory from the blue to the green region that corresponds to the agent's premature entrance into the treasure region. \\
        \textit{Failure case}: The agent enters the treasure region without collecting the key and remains there for the rest of the episode. This highlights a failure case of the model, where the hidden state ``drifts'' from the green region to the red region, i.e., the model convinces itself that the agent must have picked up the key at some point. Such \textit{causal confusion} errors are well-documented in the RM literature \cite{tien2022study}. In our case, we suspect that the error is due to a bias in our data generation method inducing a correlation between key possession and time spent in the treasure region. As described in Appendix \ref{app:dataset_generation}, we specifically screen for trajectories where the treasure is visited with the key (the ``treasure'' class) but not for those where it is visited without it. There are thus likely to be very few training trajectories that spend a lot of time in the treasure region without collecting the key, making this probe an extreme outlier on which performance is poor. Adding and selecting for a fourth ``\text{key\_no\_treasure}'' class during data generation may have mitigated this issue.
    }
    \label{fig:key_interpretability}
\end{figure}

\begin{figure}[htb!]
    \centering
    \includegraphics[width=\textwidth]{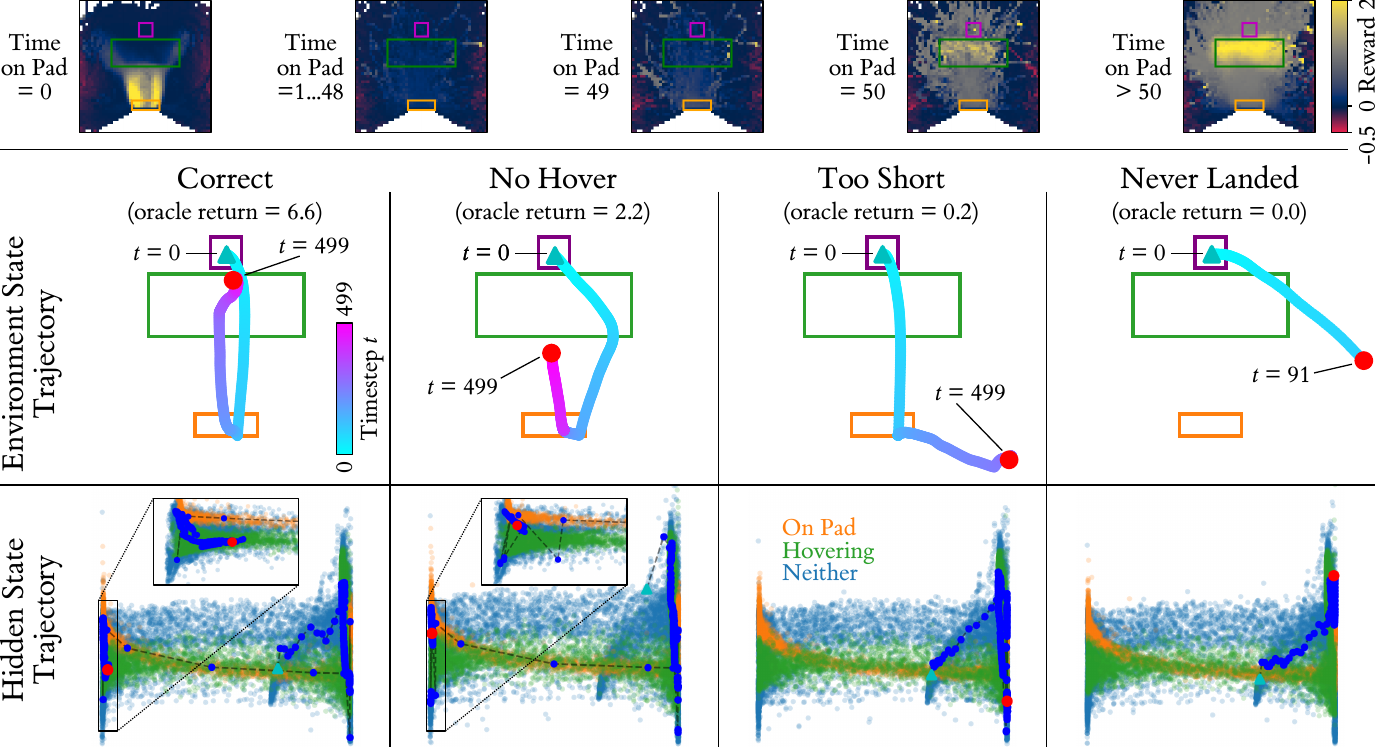}
    \caption{
        Lunar Lander task trajectory probing. \\
        \textbf{Top}: The predicted reward with respect to the agent's position and the length of time that it has been on the pad. The model has learnt to reward the agent for landing on the pad when it has not previously landed, and then reward it for taking off and remaining in the hover zone after 50 timesteps on the pad, as per the oracle. However, observe how the top half of the hover zone is attributed higher reward than the lower half. This feature is not present in the oracle (which rewards the entire hover zone equally) but makes great practical sense, as a stochastic policy is less likely to drop out of the zone under the effect of gravity if it remains some distance from the bottom edge. Also note the small negative rewards near the environment boundaries, which disincentivise positions with a high risk of leading to early termination, and the ``funnel'' of intermediate positive reward, which may help to guide the agent up from the pad to the hover zone. Collectively, these deviations from the oracle reward function could actually be seen as \textit{improvements}, and explain why we observe significantly higher performance on the $R_{\text{hover}}$ reward component when the CSC Instance Space LSTM reward model is used, compared with using the oracle itself. \\
        \textbf{Middle/bottom}: Four trajectory probes showing the model's hidden state transitions. \\
        \textit{Correct}: The agent lands on the pad as quickly as possible, waits for 50 timesteps, and then takes off again and hovers in the hover zone. The hidden state trajectory shows a transition from right to left once the agent has been on the pad for long enough. It also clearly shows the agent is in the Hovering hidden state region.  \\
        \textit{No Hover}: In this trajectory, the agent remains on the pad for too long and does not enter the hover zone. It has a similar hidden state transition to the \textit{Correct} trajectory but does not enter the Hovering hidden state region. \\
        \textit{Too Short}: The agent lands on the pad but does not stay there for the required 50 timesteps. In this particular case, the agent lands correctly but then slips out of the landing pad, coming to rest on a different area of terrain until the end of the trajectory. In the hidden state trajectory, there is no transition from the right side to the left side, matching the idea that the agent has not remained on the pad for long enough. \\
        \textit{Never Landed}: The agent does not land on the pad at all. In this particular example, the agent reaches the boundary of the environment, which causes the episode to terminate early. Similar to the \textit{Too Short} example, the hidden state trajectory does not transition from right to left.
    }
    \label{fig:lunar_lander_interpretability}
\end{figure}

In this section, we provide additional trajectory probing plots (like Figure \ref{fig:charger_interpretability}) for the Timer (Figure \ref{fig:timer_interpretability}), Moving (Figure \ref{fig:moving_interpretability}), Key (Figure \ref{fig:key_interpretability}), and Lunar Lander (Figure \ref{fig:lunar_lander_interpretability}) tasks. As before, in these probing plots, we analyse the best-performing CSC Instance Space LSTM model for each task (according to the reward reconstruction metric).

\vfill

\vfill

% \clearpage
% % \newpage
% {\fontsize{9}{0}\bibliography{bibliography.bib}}

\end{document}